\documentclass[a4paper]{article}
\title{Symbolic integration by integrating learning models with different strengths and weaknesses}
\author{Hazumi Kubota$^1$, Yuta Tokuoka$^1$, Takahiro G. Yamada$^{1,2}$ \& Akira Funahashi$^{1,2,*}$}
\date{}
\usepackage{diagbox}
\usepackage[pdftex]{graphicx}
\usepackage{url}
\usepackage{a4wide}

\newenvironment{affiliations}{%
    \setcounter{enumi}{1}%
    \setlength{\parindent}{0in}%
    \slshape\sloppy%
    \begin{list}{\upshape$^{\arabic{enumi}}$}{%
        \usecounter{enumi}%
        \setlength{\leftmargin}{0in}%
        \setlength{\topsep}{0in}%
        \setlength{\labelsep}{0in}%
        \setlength{\labelwidth}{0in}%
        \setlength{\listparindent}{0in}%
        \setlength{\itemsep}{0ex}%
        \setlength{\parsep}{0in}%
        }
    }{\end{list}\par\vspace{12pt}}

\begin{document}
\maketitle
\begin{affiliations}
 \item Center for Biosciences and Informatics, Graduate School of Fundamental Science and Technology, Keio University, Kanagawa, Japan.
 \item Department of Biosciences and Informatics, Keio University, Yokohama, Kanagawa, Japan.
 \item[*] funa@bio.keio.ac.jp
\end{affiliations}
\section*{Abstract}
Integration is indispensable, not only in mathematics, but also in a wide range of other fields.
 A deep learning method has recently been developed and shown to be capable of integrating mathematical functions that could not previously be integrated on a computer.
However, that method treats integration as equivalent to natural language translation and does not reflect mathematical information.
In this study, we adjusted the learning model to take mathematical information into account and developed a wide range of learning models that learn the order of numerical operations more robustly.
In this way, we achieved a 98.80\% correct answer rate with symbolic integration, a higher rate than that of any existing method. We judged the correctness of the integration based on whether the derivative of the primitive function was consistent with the integrand. By building an integrated model based on this strategy, we achieved a 99.79\% rate of correct answers with symbolic integration.

\section{Introduction}
 Integration is required in simulations to predict the behavior of objects in ﬁelds such as computer science, aeronautics, mechanical engineering, and systems biology \cite{moses1974macsyma,lynch2001effects,fiorentini2009nonlinear,goldbeter1991minimal}.
 Traditionally, numerical integration has been carried out on computers.
 Various methods of numerical integration have been proposed to approximate small changes of any mathematical function (the integrand) to be integrated.
 The approximation of small changes, however, always leads to errors.
 Furthermore, numerical integration requires an initial value of the integrand, i.e., it obtains only particular solutions.
 A resolution of these problems and limitations is symbolic integration, by which an exact, general solution can be derived.
 In computerized symbolic integration, the integrand is transformed via a theorem to obtain an integrated function, which is called a primitive function.
 This symbolic integration process, formulated by Robert Risch in 1970 \cite{risch1970}, is known as Risch's algorithm, and software based on this algorithm, Maxima (\url{http://maxima.sourceforge.net/}), has been developed.
 However, Maxima relies on approximations because Risch's algorithm involves finding roots, and there is currently no method to derive an analytical expression for a root.
 Some integrands, therefore, cannot be integrated.
 Programs have been developed for Mathematica (\url{https://wolfram.com/mathematica/}) and Matlab (\url{https://mathworks.com/products/matlab/}) as well as by Rubi \cite{Rich2018} to perform symbolic integration, but even these programs cannot integrate some functions.

 In contrast, deep learning has been widely used in natural language processing to formulate algorithms for language translation \cite{liang2016neural,devlin2018bert}.
 Deep-learning models such as Recurrent Neural Network (RNN) and Transformer have been shown to be Turing-complete and are expected to facilitate learning arbitrary algorithms \cite{perez2019turing}.
 Saxton et al. have successfully used Long Short-Term Memory (LSTM), a type of RNN, and the Transformer model to solve a wide range of mathematical problems, including arithmetic operations and mathematical simplification \cite{saxton2019analysing}.
 Lample et al. \cite{Lample2020Deep} have constructed a Transformer model with the integrand as input and a primitive function as output and have shown that the trained model can solve some problems that could not be solved by existing mathematical integration software.
 However, Lample et al. \cite{Lample2020Deep} limited the range of constants in the functions used to train the Transformer model, and the generation of the functions by sampling resulted in a lack of coverage of training data.
 In the Transformer model, the mathematical expression of the integrands in Polish notation makes it difficult to learn the syntactic structure of mathematical expressions because the operators and their subjects, which should be close to each other to clarify the order of evaluation of mathematical operations, are far from each other.
 For example, if the integrand is $(a+b)\times c$, the Polish notation would be ``$\times\ +\ a\ b\ c$'', where ``$c$'' is the distal part of the ``$\times$'' operation.
 The syntactic structure of the expression is thus difficult to decipher.
 Furthermore, the Transformer model adopted by Lample et al. \cite{Lample2020Deep} learns the exact positions of mathematical expressions as input; however, recent research has shown that it is useful to consider the relative positions as input when solving problems related to conversion of variable output sequences from variable input sequences, such as machine translation \cite{huang-etal-2020-improve,shaw-etal-2018-self}.
 In addition, a property in the problem setting of symbolic integration allows verification of whether the primitive function output of the learning model agrees with the integrand when differentiated.
 It is thus possible to construct a highly accurate symbolic integration algorithm by using a variety of input formats and learning models and adopting the output of the learning model that answers correctly.

\begin{figure}[ht]
 \begin{center}
  \includegraphics[width=15cm]{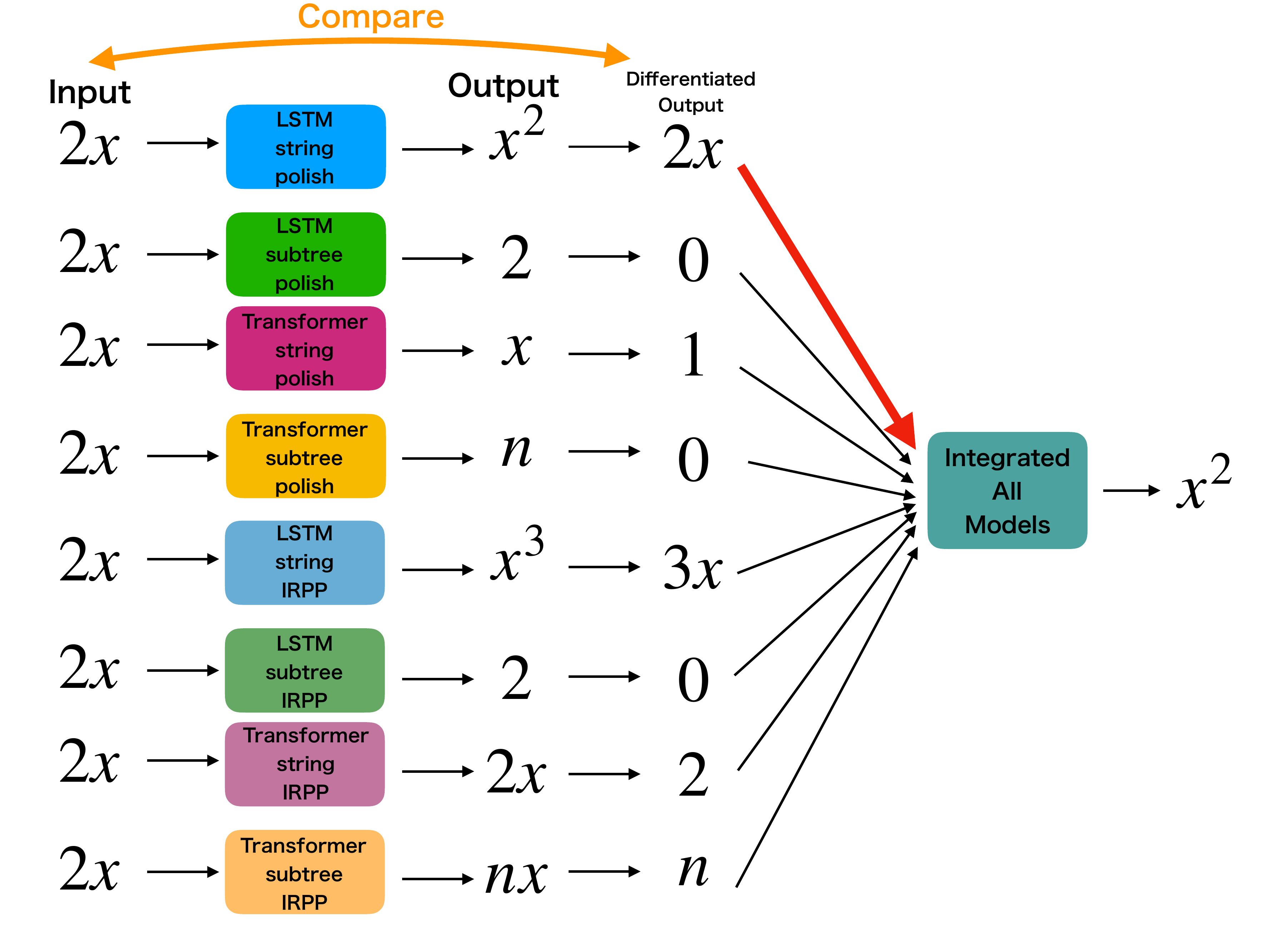}
  \caption{Schematic of the Integrated All models for learning symbolic
  integration developed in this study.  A learning model based on Long
  Short-Term Memory (LSTM) and Transformer was developed as a learning
  model that took an integrand as input and outputted a primitive
  function. As input-output formats of these learning models, we
  constructed a method in which input-output was a string of the same
  functions as the learning model developed by Lample et
  al. \cite{Lample2020Deep} (string). The input-output was a subtree of
  an Abstract Syntax Tree (AST) that described the functions proposed in
  the study (subtree). We developed a scheme in which an integrand and a
  primitive function were described in Polish notation (polish), a
  scheme in which an integrand was described in reverse Polish notation,
  and a primitive function that was described in Polish notation
  (Integrand Reverse polish Primitive Polish, IRPP). Eight learning
  models were developed. Symbolic integration was finally performed by
  choosing the result that satisfied the requirement that the
  differentiated function matched the input integrand.}
  \label{fig:integrated_all_model}
 \end{center}
\end{figure}

 To address these problems, we created a new dataset consisting of functions created by comprehensively multiplying elementary functions.
 For the input and output of the learning model, we used a subtree method that represented robustly the relationship between the operator and its subject in the arithmetic expression of the function.
 We also used a new Integrand Reverse polish Primitive Polish (IRPP) scheme, in which the integrands are input in reverse Polish notation, and the primitive functions are output in Polish notation.
 We constructed an LSTM that learned the relative positions of inputs better than the Transformer model.
 Finally, we developed a symbolic integration algorithm that used the correct model among eight learning models trained with different input-output schemes (Figure \ref{fig:integrated_all_model}).
 When Polish notation was used as the input-output scheme and LSTM as the learning model, the output primitive function was correct 98.80\% of the time.
 This percentage was higher than that of all the existing symbolic integration methods, including Lample et al.'s method \cite{Lample2020Deep}.
 By integrating the learning models, we established a symbolic integration method that increased the correct answer percentage by $\sim$1.0\% to 99.79\%, which exceeded the improvement (0.5\%) of the correct answer rate achieved with any single learning model and with the runner-up, Mathematica (98.30\%).
\section{Results}
\subsection{Evaluation of symbolic integration methods}
 In this study, we created 12,122 new pairing functions of integrand and primitive functions.
 Using these functions as training data, we constructed a learning model consisting of an LSTM and a Transformer with integrands as input and primitive functions as output.
 The input-output format of the functions was adjusted to the format of Lample et al. \cite{Lample2020Deep} (string) or a combination of subtrees consisting of three nodes in the Abstract Syntax Tree (AST) (subtree).
 The order of mathematical expressions was adjusted to both Polish notation (polish) or reverse Polish notation for the integrand and to Polish notation for the primitive function (IRPP).
 Eight learning models were constructed by adjusting the learning model, the input-output scheme as a string/subtree, or the order of the mathematical expression as polish/IRPP (see details in section \ref{material:createData}).
 Of the functions created, 80\% were split into training data (9,697 functions), and 20\% into test data (2,425 functions).
 Each model was then trained using 10-fold cross-validation on the training data.
 For these trained models, we used the integrands of the test data as input and evaluated the rate of correct answers, i.e., the percentage of matches between the derivative of the output primitive function and the integrand (Figure \ref{fig:accuracy_result}).
 The LSTM model with string and polish showed the highest correct answer rate as a stand-alone learning model (98.80\%).

 To assess the superiority of this learning model for symbolic integration, we calculated the correct answer rate for 2,425 test functions in mathematical integration tools such as Mathematica, Maxima, Rubi, and Matlab (Figure \ref{fig:accuracy_result}).
 We found that LSTM with string and polish outperformed Mathematica (98.30\%), which achieved the highest correct answer rate of any previous symbolic integration tool.
\subsection{Improvement based on integration of the learning model}
 The correctness of the algorithm for symbolic integration can be veriﬁed by comparing the derivative of the output primitive function with the integrand.
 We constructed a model that generated a primitive function, the derivative of which consistently equaled the integrand for the eight learning models constructed in this study (Figure \ref{fig:integrated_all_model}, Integrated All models).
 The correct answer rate of the Integrated All models on the test data was 99.79\%, an improvement of 0.99\% (24 functions) over the highest accuracy of the stand-alone models.
 Mathematica's correct answer rate on test data was 98.30\%, and the improvement of the best stand-alone learning model was 0.5\% over Mathematica.
 The result with the Integrated All models was an even more remarkable improvement.
\subsection{Reason for integrating learning models to improve accuracy}
Integrating the output of the learning models that answered correctly produced a significant improvement in the accuracy of the Integrated All models for the test data.
 The main reason for this improvement was the significant differences in the correctness of the answers between learning models.
 We determined the number of integrands that were calculated incorrectly among the symbolic integration results with test data (Figure \ref{fig:venns}).
 We confirmed that the integrands that were incorrectly calculated depended very much on the learning model (LSTM/Transformer) and input-output schemes of the datasets (string/subtree and polish/IRPP) (Figure \ref{fig:venns}a, b).
 Because the number of incorrectly calculated integrands for any input-output scheme or learning model was no greater than five (Figure \ref{fig:venns}c), the Integrated All models allowed the adoption of correctly answered questions and avoided incorrect answers.
\subsection{Characterization of the symbolic integration mechanism}
 The reason that the integration of the output results by the learning models improved the accuracy was that each learning model could answer different integrands correctly; each learning model performed the symbolic integrations based on different symbolic integration mechanisms.
 To elucidate these differences, we analyzed the attention layer of each learning model, which represents the part of the integrand and the primitive function that each learning model focuses on.

 For the LSTM-based learning model, we focused on the attention map when the primitive functions were $\frac{\cos^3 x}{x}$ and $\frac{\cos^4 x}{x}$ in the learning model with string and polish, and in the learning model with subtree and polish.
 The functions where the primitive functions were $\frac{\cos^3 x}{x}$ and $\frac{\cos^4 x}{x}$ were generalized to the problem of turning the integrand $-\frac{\cos^{-1+n} (x)(\cos (x)+n x \sin (x))}{x^{2}}$ into the primitive function $\frac{\cos^n x}{x}$.
 In other words, the symbolic integration of this integrand is a problem that can be solved based on the formula with attention to the constant $n$.
 By measuring the similarity of the attention map generated from the input of these two functions, we could verify whether LSTM was able to evaluate the integrals in accordance with the integration formula.
 Visualization of the attention maps made it clear that the attention maps were similar to each other in both learning models with string and polish as well as with subtree and polish (Figure \ref{fig:LSTM_attention}a, b, d, e).
 However, a comparison of the LSTM's attention map with string and polish based on the Jensen-Shannon divergence (JS divergence) \cite{JSDivergence} revealed that the JS divergence of the attention map for the constant 3, 4 was much larger than the JS divergence for the other expressions (Figure \ref{fig:LSTM_attention}c).
 The implication was that the learning model could perform the symbolic integration based on the integration formula while giving appropriate attention to the constant $n$.
 A comparison of the LSTM's attention maps with subtree and polish using JS divergence showed that in addition to the JS divergence of the attention map of the different subtrees in the primitive function (between ``power cos 4'' and ``power cos 3'' subtrees, and ``4 EOS (End of Sentence) EOS'' and ``3 EOS EOS'' subtrees), the JS divergence of the attention map of the common subtrees (between the ``divide power x'' subtree for these two integrands and the ``cos x EOS'' subtree for these two integrands) was much larger than that of other subtrees (Figure \ref{fig:LSTM_attention}f).
 The implication was that in addition to gazing at the constant $n$ based on the integration formula, the learning model was able to perform the mathematical integrals of gazing that went beyond the integration formulas.
 These results revealed that the learning models based on LSTM learned different methods of symbolic integration from each other because of the diﬀerences in input-output schemes and that each model produced different results in terms of the correctness or incorrectness of the primitive function.

 Next, for the Transformer-based learning model, we focused on the self-attention layer of the encoder when the integrand was $e^x\times n \times \cos^3{x}(\cos {x} - 4 \sin {x})$ in the learning model with string and polish as well as subtree and polish.
 The Transformer model consisted of six self-attention layers for each of the encoder and decoder parts, and there were eight multi-heads in each layer.
 We attempted to clarify the symbolic integration mechanism of the Transformer model by uncovering the part of the integrand on which the model focused for each head in this self-attention layer and for each layer.
 To clarify the relationship between the gazed part of the integrand in each head, we mapped the attention maps of the heads to the two-dimensional space using the multidimensional scaling method based on the JS divergence between the attention maps of the 48 heads (Figure \ref{Transformer_embeded_2D}).
 The results showed that the attention maps of the heads in the same layer tended to become similar in deeper layers, regardless of whether the string and polish or the subtree and polish were used (Figure \ref{Transformer_embeded_2D}a, c).
 Regardless of the input-output schemes, the Transformer model paid attention only to the part of the function to which it decided to pay attention in each deep layer of the self-attention layer.
 In the shallow layers, however, each head apparently had a very diﬀerent attention map for both input-output schemes.
 The implication is that, depending on the input-output schemes, the Transformer model focused on each unique part of the integrand on a shallow layer, i.e., various parts of the integrand for each head in the initial processing stage of the symbolic integration.
 To further assess the similarity of the attention maps between the heads in each of these layers, we calculated the average entropy of the attention maps of the eight heads in each layer (Figure \ref{fig:Entropies of attention distributions}).
 The result showed that the average entropy, which is the homogeneity between attention maps, was low in the shallow layers, but an increase of the average entropy in the deeper layers indicated that the diﬀerence in the attention maps between the heads had disappeared.
 The more pronounced increase of average entropy with each layer for the subtree and polish input-output schemes (Figure \ref{fig:Entropies of attention distributions}a) than for the string and polish input-output schemes (Figure \ref{fig:Entropies of attention distributions}b) indicated that the model with string and polish speciﬁed the parts of the integrand on which to concentrate from the beginning of the input of the integrand throughout the head.
 In contrast, the model with subtree and polish concentrated on a part of the integrand specific to each head and performed symbolic integration based on a wide range of information.
 In the case of the string and polish input-output schemes, the attention maps of each head were similar to each other (Figure \ref{Transformer_embeded_2D}c), but in the case of the subtree and polish input-output schemes, the attention maps of each head were different (Figure \ref{Transformer_embeded_2D}d).
 In other words, in the case of the string and polish, the role of each head was firmly fixed to the part of the integrand that was the focus of attention across the layers in the self-attention layer, but in the case of the subtree and polish, the part of the integrand on which attention was focused was more flexible and was determined for each head without being bound by the relationship between the heads of the layers.

There were thus differences in the (i) relationships between the paired mathematical expressions that depended on the input-output schemes and the learning models , (ii) mathematical expressions of the integrand, and (iii) processing methods in the model.
 Attention should be paid to these differences in the conversion from the integrand to the primitive function because the differences they created caused the functions to be integrated to be correct or incorrect.
 The integration of the output results of the learning models consequently improved the correct answer rate.
\subsection{Evaluation of the execution time for symbolic integration}
 To determine whether the execution times of symbolic integration for the eight learning models were realistic, we measured the average runtime for a single integration of the test data using these models and existing integration tools (Supplementary Figure 1).
 The average execution time for integration varied from 0.062 seconds (LSTM with string and polish) to 1.059 seconds (Transformer with subtree and polish) and was shorter than the time required for the existing symbolic integration tools, except for Matlab.
\section{Discussion}
\subsection{Optimal input-output schemes and learning model}
\begin{figure}[ht]
 \begin{center}
  \includegraphics[width=15cm]{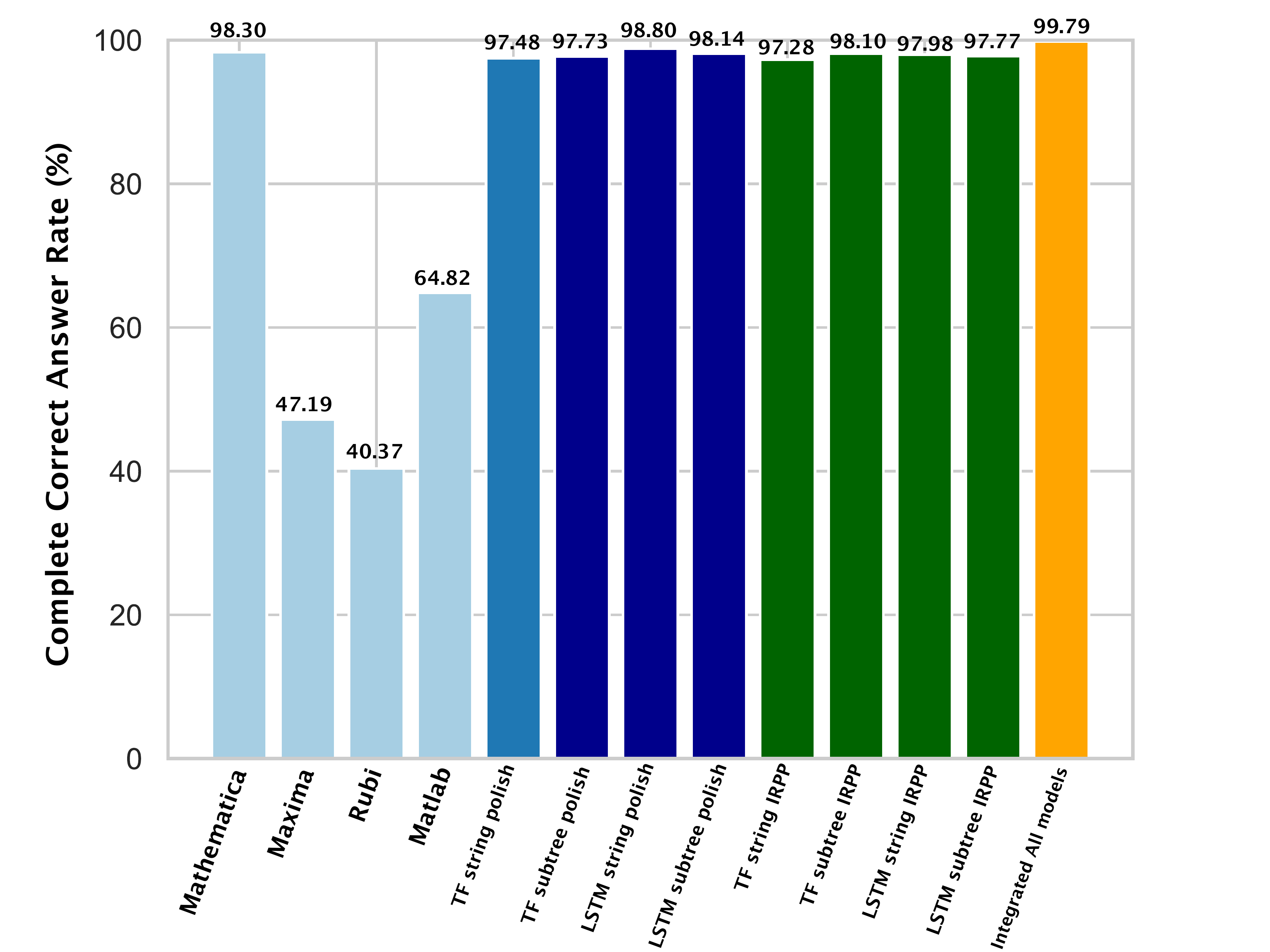}
  \caption{Rate of correct answers to test data for existing symbolic
  integration tools and the learning model developed in this study.  The
  rate of correct answers (i.e., the percentage of derivatives of output
  primitive functions that matched the input integrands) is shown for
  existing symbolic integration tools (light blue, Mathematica, Maxima,
  Rubi, Matlab), the deep learning model developed by Lample et
  al. \cite{Lample2020Deep} (blue, Transformer string polish, TF string
  polish), the learning models developed in this study (dark blue and
  green), and the Integrated All models (orange).}
  \label{fig:accuracy_result}
 \end{center}
\end{figure}

\begin{figure}[ht]
 \begin{center}
  \includegraphics[width=15cm]{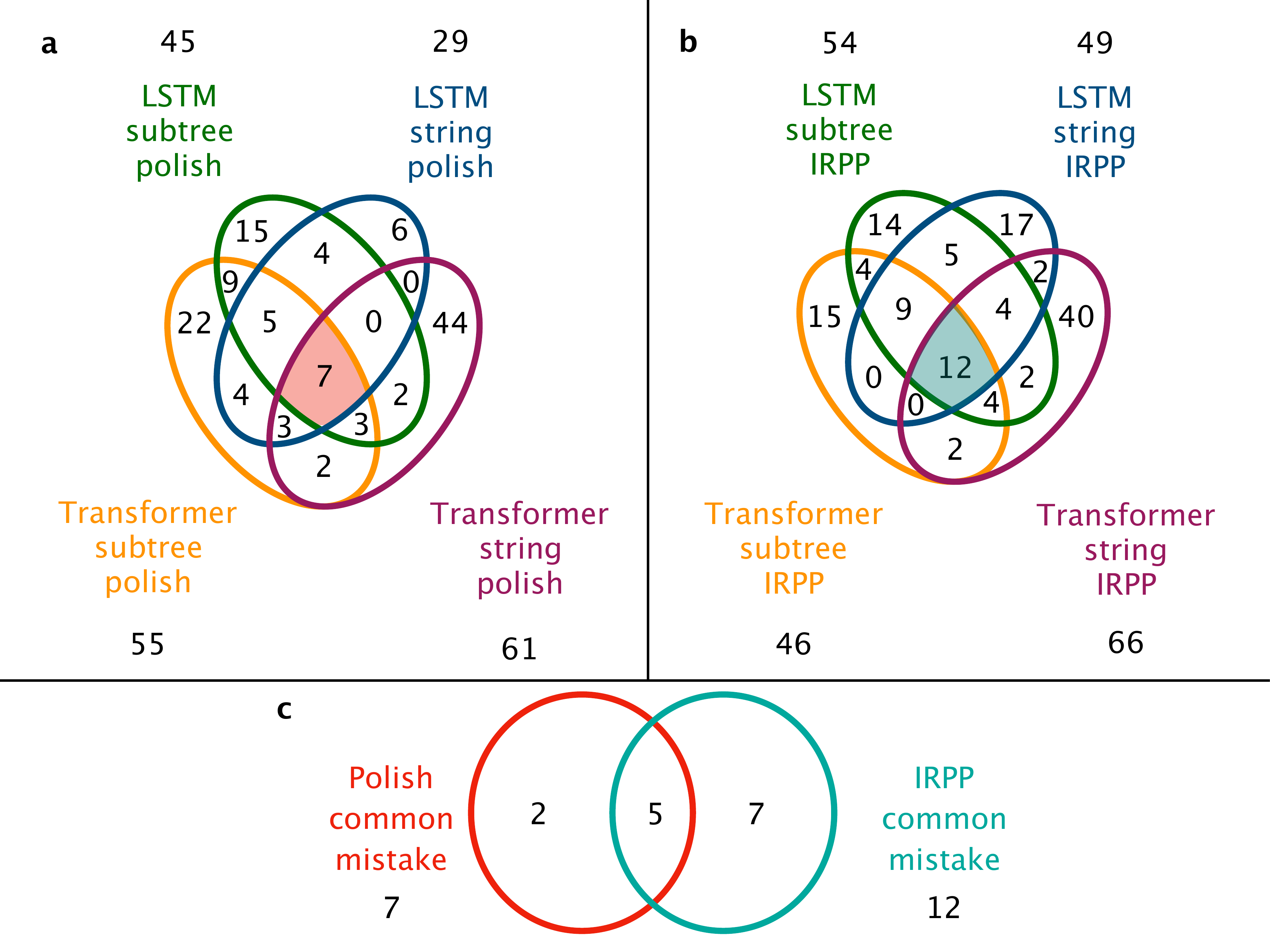}
  \caption{Venn diagrams of the numbers of incorrectly answered
  integrands for the test data by each model.  Venn diagrams of the
  numbers of incorrectly answered integrands for test data when using
  the string and subtree formats for Long Short-Term Memory (LSTM) and
  Transformer. {\bf a}: polish input-output scheme and {\bf b}: IRPP
  input-output scheme. {\bf c}: Venn diagram of the number of
  incorrectly answered integrands to the test data for the learning
  model with all-polish input-output schemes and with all-IRPP
  input-output schemes. The number near the name of each learning model
  indicates the number of incorrectly answered integrands for the test
  data when the designated model was used.}
  \label{fig:venns}
 \end{center}
\end{figure}

\begin{figure}[ht]
 \begin{center}
  \includegraphics[width=15cm]{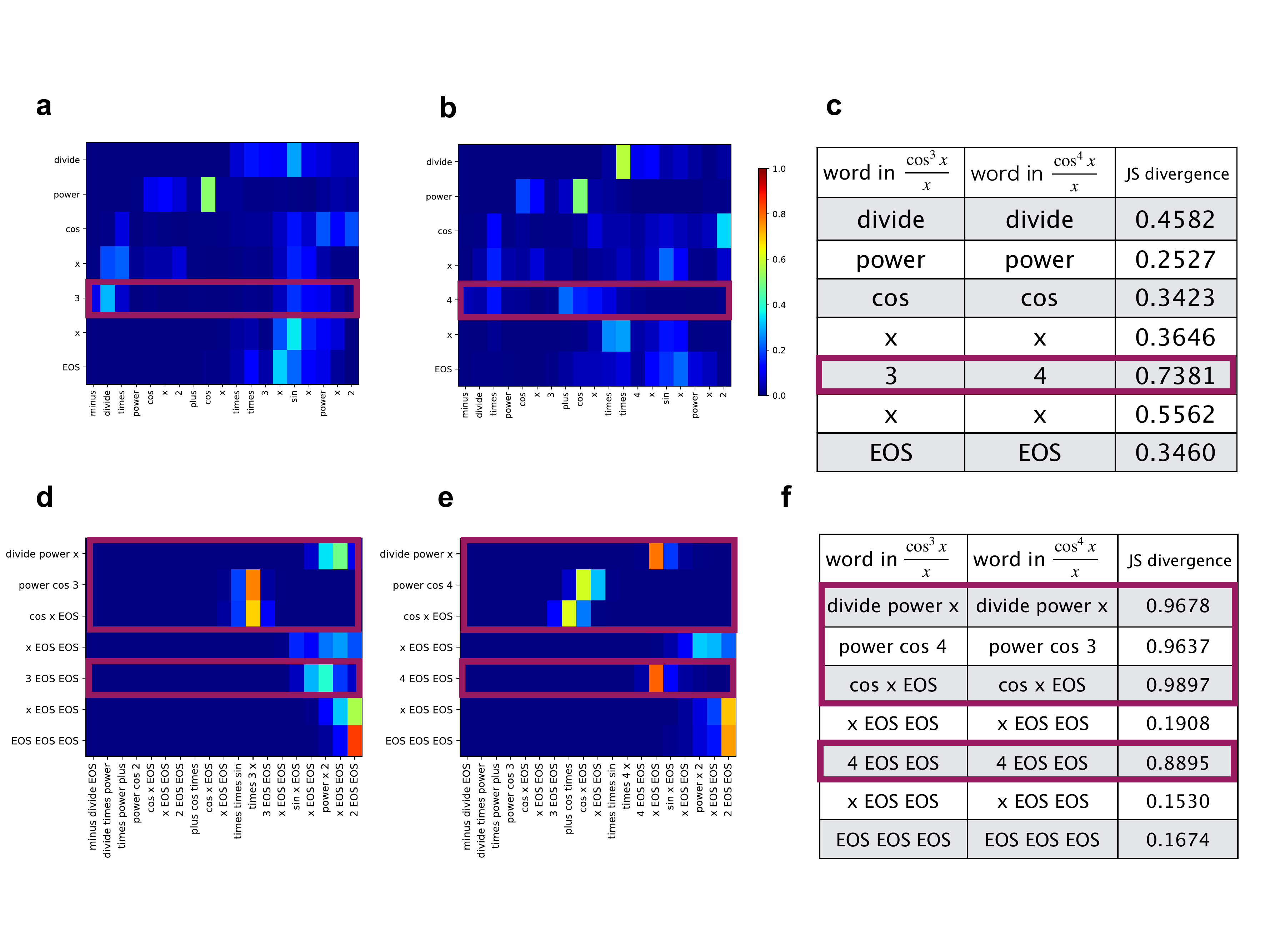}
  \caption{Visualization of the attention map of the LSTM model. {\bf a,
  b, d, e}: Visualization of the attention map of the LSTM. Each element
  in a heat map is the weight of the LSTM connecting the integrand and
  the primitive function to the LSTM for each input or output token or
  subtree. {\bf a}: The result of using an integrand whose primitive
  function was $\frac{\cos^3 x}{x}$ as input to the LSTM with string and
  polish schemes. {\bf b}: The result of using an integrand whose
  primitive function was $\frac{\cos^4 x}{x}$ as input to the LSTM with
  string and polish schemes. {\bf d}: The result of using an integrand
  whose primitive function was $\frac{\cos^3 x}{x}$ as input to the LSTM
  with subtree and polish schemes. {\bf e}: The result of using an
  integrand whose primitive function was $\frac{\cos^4 x}{x}$ as input
  to the LSTM with subtree and polish schemes. {\bf c,f}: The
  Jensen-Shannon divergence in the LSTM attention map when outputting
  $\frac{\cos^3 x}{x}$ and $\frac{\cos^4 x}{x}$ as primitive functions
  was calculated for each token or subtree. {\bf c}: LSTM with string
  and polish schemes. {\bf f}: LSTM with subtree and polish schemes.}
  \label{fig:LSTM_attention}
 \end{center}
\end{figure}

\begin{figure}[ht]
 \begin{center}
  \includegraphics[width=15cm]{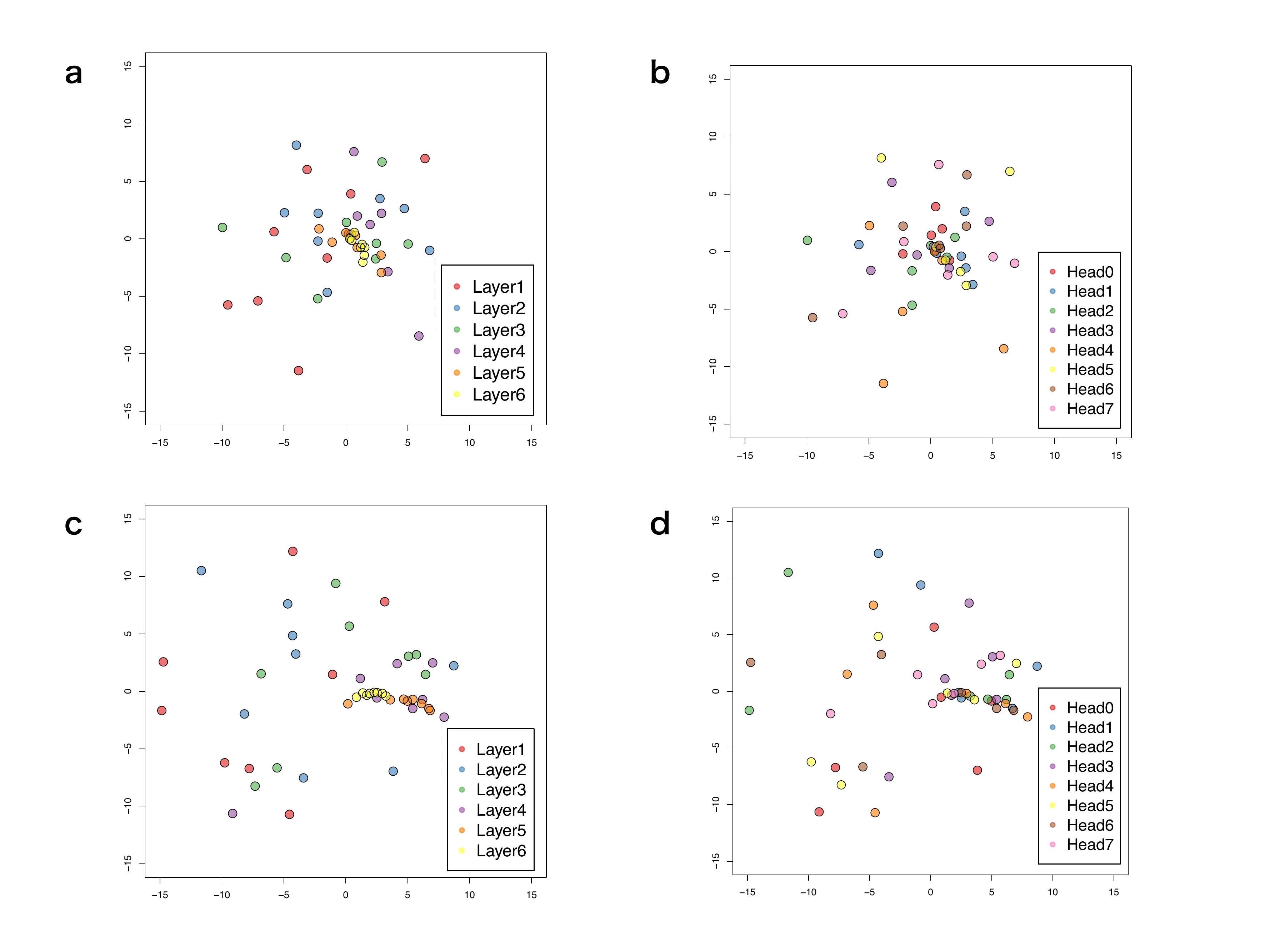}
  \caption{Relationship between the attention maps of each head in the
  self-attention layer of the Transformer model. The attention map of
  each head in the self-attention layer of the Transformer model when
  $e^x\times n\times \cos^3{x}(\cos {x} - 4\sin {x})$ was input as the
  integrand was plotted in two-dimensional space using the
  multidimensional scaling method with Jensen-Shannon divergence as the
  dissimilarity {\bf a,b}: for the Transformer model using the string
  and polish schemes {\bf a}: shown in the same color for each layer,
  and {\bf b}: shown in the same color for each head, {\bf c,d}: for the
  Transformer model using the subtree and polish schemes {\bf c}: shown
  in the same color for each layer, and {\bf d}: shown in the same color
  for each head.}
  \label{Transformer_embeded_2D}
 \end{center}
\end{figure}

\begin{figure}[ht]
 \begin{center}
  \includegraphics[width=15cm]{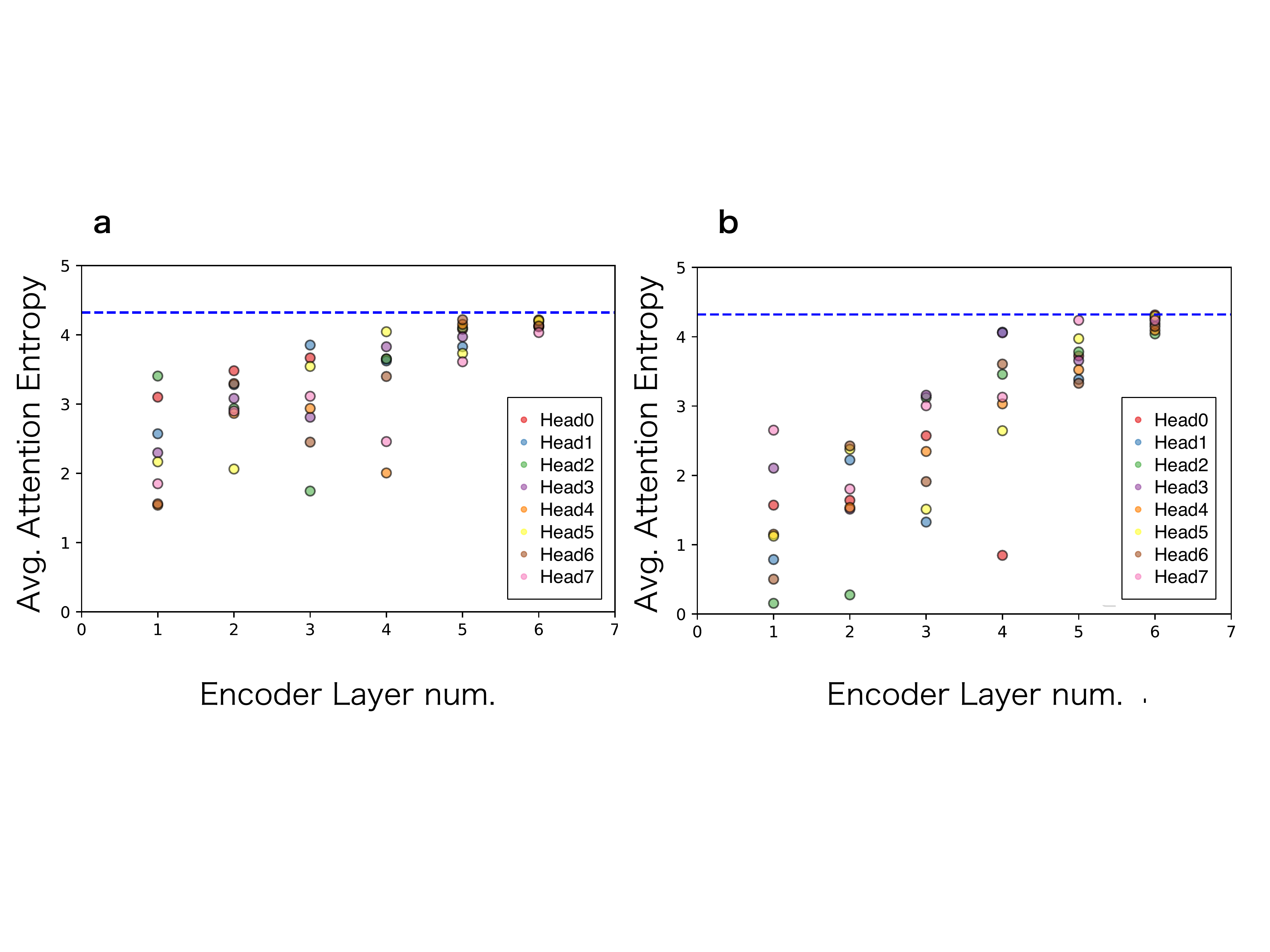}
  \caption{Transition of the average entropy of the attention map in the
  head as the layers were deepened in the self-attention layer of the
  Transformer model.  The average entropy of the attention map at each
  head of each layer in the self-attention of the Transformer model when
  $e^x\times n\times \cos^3{x}(\cos {x} - 4\sin {x})$ was input as the
  integrand for the Transformer model {\bf a}: using the string and
  polish schemes, and {\bf b}: using the subtree and polish schemes. The
  blue dotted lines show the average entropy when the attention was
  uniform for all tokens or subtrees.}
  \label{fig:Entropies of attention distributions}
 \end{center}
\end{figure}

 In this study, we pointed out various problems in the learning model for symbolic integration developed by Lample et al. \cite{Lample2020Deep}
 We developed eight learning models based on combinations of each input-output scheme or learning model, and we showed that the learning model with string and polish and LSTM achieved the highest complete correct answer rate of all learning models and existing symbolic integration tools (Figure \ref{fig:accuracy_result}, 98.80\%, LSTM string polish).

 The main difference between this model and that of Lample et al. \cite{Lample2020Deep} was the use of LSTM instead of Transformer as the learning model.
 The superior performance of LSTM reflected the difference in the way LSTM and Transformer learn the location of the tokens in the input integrands.
  The LSTM model learns the relative positions of the tokens in the input integrands as it sequentially updates the memory cells inside LSTM by sequentially feeding the input tokens as time-series data into the model.
 In contrast, the Transformer adds positional information about the order of the tokens of the integrands before they are input to the model.
 In addition, the biased length of one of the functions in the dataset used in this study (Supplementary Figure 2) caused the learning model to take a variety of long and short functions as input and processed them appropriately to convert them into primitive functions.
 When the input consists of such a variety of long and short functions, the use of absolute positioning of mathematical symbols is detrimental.
 For example, if the learning model is good at integrating over short functions, it will have to integrate over long functions by anticipating unpredictable information about the positions of function symbols.
 If it is good at integrating over long functions, it will fail to integrate over short functions because it will not be able to use enough information about the positions of function symbols.
 However, if a learning model is based on the relative positions of mathematical symbols, these adverse effects apparently do not exist, and for this reason LSTM outperforms Transformer.
 It was thus useful to adopt a learning model that could learn the order of the tokens in the input functions as relative positions in the symbolic integration.
 This utility was confirmed by the fact that the proposed subtree input-output schemes improved the accuracy of the Transformer model compared to the string input-output scheme by considering the relative input order (Figure \ref{fig:accuracy_result}, 97.48\%, Transformer string polish, 97.73\%, Transformer subtree polish).

 It is interesting to note, however, that the accuracy was lower when the subtree input-output scheme was used with LSTM than when the string scheme was used (Figure \ref{fig:accuracy_result}, 98.14\%, LSTM subtree polish), probably because the attention from the decoder part to the output of the encoder part of the model in LSTM  is performed only once for each token in the input function and only for one context.
 In addition, the IRPP input-output scheme, which tries to make the output of the symbolic integration consider the rules of operation, did not improve the accuracy of the LSTM symbolic integration with string input (Figure \ref{fig:accuracy_result}, 97.98\%, LSTM string IRPP).
 We considered that this failure was due mainly to the use of different ways to describe the inputs and outputs, which are both homogeneous in terms of functions.
 The effort required to learn the different rules of arithmetic consequently increases, and this increased effort leads to more errors in symbolic integration.
 If the implications of the input and output are essentially homogeneous, as in symbolic integration, then the improved accuracy could probably be attributed to the unification of notation.
\subsection{Affinity of input-output schemes and learning models}
In this study, we achieved a 99.79\% correct answer rate with an integrated model that chose the output of the model that answered correctly.
 This improvement was achieved because there was a difference in the integrands that each learning model could correctly integrate (Figure \ref{fig:venns}).
 To explore this issue in more depth, we tried to elucidate the symbolic integration of the integrands that each learning model incorrectly integrated by focusing on the integrands that were integrated incorrectly by each learning model.

 We aimed to elucidate the characteristics of integrands that were incorrectly integrated by focusing on the wrong answers caused by the difference between string and subtree input-output schemes.
 We investigated the functions that were incorrectly answered by the four learning models with string and the four learning models with subtree (Supplementary Table 1).
 The results confirmed that the learning model with string tended to produce results contrary to the  rules of mathematical operation grammar.
 In those results, extra mathematical symbols were present or some mathematical symbols were missing when the input was an integrand composed of long mathematical expressions (Supplementary Table 2).
 The missing symbols were probably due to the fact that the operators and operands of the arithmetic relationships were located far from each other, and it was consequently impossible to fully grasp the arithmetic rules.
 We confirmed that in the learning model with subtree, only the symbol of one node in the subtree was wrong in the output primitive function when the input was an integrand composed of short mathematical expressions (Supplementary Table 3).
 This result likely reflected the requirement of the learning model with subtree for the output of a large amount of information from a small amount of information when integrating the functions, because its output was the mathematical symbols of the three nodes of the subtree from one hidden vector propagated from the input.
 This lack of information may have led to errors in the mathematical symbols of only one node of the subtree.

 By focusing on the difference between LSTM and Transformer in terms of wrong answers, we hoped to identify the characteristics of the integrands that were incorrectly integrated.
 We investigated the functions that were incorrectly integrated by the four models that adopted LSTM and by the four models that adopted Transformer (Supplementary Table 4).
  We confirmed in the case of the LSTM-based learning model that the model evaluated the integral incorrectly when the lengths of the integrand and the primitive function were similar (Supplementary Table 5).
 This pattern resulted from the fact that the propagation of information inherited from the encoder was insufficient for the output of a primitive function of similar length, and consequently the information added by the attention layer could not be used effectively.
 In the case of the Transformer-based learning model, the answers tended to be wrong when the length of the integrand and the primitive function were significantly different (Supplementary Table 6).
 In the Transformer model, the same number of self-attentions were set in the encoder and decoder part, and when the information contents of the input and output functions were similar, the encoding and decoding worked effectively.
 However, when the information content differed significantly, wrong answers were produced because of the imbalance in the accuracy of the encoding and decoding.

 We aimed to clarify the characteristics of integrands that were incorrectly integrated by focusing on the problem of wrong answers due to the diﬀerence between the polish and IRPP input-output schemes.
 We investigated the number of wrong answers in four learning models with polish and four learning models with IRPP (Supplementary Table 7).
 The learning models with polish tended to produce incorrect answers for functions with few operators (Supplementary Table 8).
 The use of Polish notation for both the input and output functions meant that the operators in the input integrands and in the relevant output primitive function were situated distally.
 The high probability that the few operators in the integrand and primitive function were strongly related to each other suggested that use of the polish input-output scheme to conﬁgure these operators distally was detrimental and led to incorrect integrands.
 We confirmed, however, that the large number of operators in the learning model with IRPP caused erroneous answers, especially in functions containing a large number of constants (Supplementary Table 9).
 In the IRPP input-output scheme, the integrand and primitive function were written in reverse Polish notation and Polish notation.
 The initial parts of both functions were thus proximal to each other.
 However, if the function was a relatively long mathematical expression with many operators, there would be many constants, and the input-output relationship of these operators would be distal.
 This condition may have been detrimental and have led to erroneous answers when the learning model processed functions composed of numerous operators and constants.
 
 In summary, it was clear that the symbolic integrations that were problematic differed among the different input-output schemes and learning models.
 The integrated model took these considerations into account and selected results in a cross-subsidized manner that led to dramatically improved accuracy.
\subsection{Usefulness of elucidating symbolic integration mechanisms}
In addition to these analyses of wrong answers, we also conducted analyses focused on the attention layer of each learning model to elucidate the mechanism of symbolic integration speciﬁc to each model and to clarify differences between them.
 We found that LSTM with string and polish was good at performing formula-compliant integrals, and the Transformer with string and polish could precisely determine the roles played by functions of interest for each head.
 These findings could not only contribute to the discovery of new  symbolic integration theorems but also facilitate the construction of more accurate symbolic integration learning models.
 For example, the performance of formula-compliant symbolic integration by LSTM with string and polish made it possible to input an unsolved integrand and analyze the attention layer to clarify the variables and mathematical symbols of interest.
 The knowledge forthcoming from this discovery should greatly facilitate the application of symbolic integration theory.
 The discovery that the role of each head in the Transformer with string and polish facilitated integration of mathematical expressions could enable improvement of the learning model via adoption of information about the substructure of the essential mathematical expression in a way that promotes this mechanism.
 We were thus convinced that characterization of learning models provides the seeds for new mathematical knowledge and very useful insights into how to devise better learning models.
\section{Conclusions}
 We developed eight learning models for symbolic integration that were combinations of input-output schemes (string/subtree, polish/IRPP) and learning models (LSTM/Transformer).
 Among these models, the learning model that adopted string and polish as input-output schemes and LSTM as its learning model achieved the highest correct answer rate (98.80\%) based on the agreement between the derivative of the output primitive function and the integrands that were not involved in the learning process.
 This learning model surpassed all existing symbolic integration methods.
 If the metric of the accuracy of symbolic integration is the agreement between the derivative of the output primitive function and the input integrand, we succeeded in constructing a symbolic integration algorithm that raised the correct answer rate of the stand-alone learning model by $\sim$1.0\% to 99.79\%.
\section{Methods}
\subsection{Creation of a dataset of integrands and primitive functions} \label{material:createData}
 To comprehensively generate data for pairs of integrands and primitive functions, we created primitive functions by multiplying up to five elementary functions chosen from $x, n$ (constant), $\sin x, \cos x, \tan x, \log (x), \exp (x), \sqrt{x}, \sqrt[3]{x}$.
 We then differentiated the created primitive function using the $D$ function in Mathematica to obtain the integrand paired with the primitive function.
 Finally, for the obtained pairs of integrands and primitive functions, we used Mathematica's $simplify$ function to simplify the functions and remove duplicates.
 The above procedure created 12,122 independent functions as pairs of integrands and primitive functions.
 The 12,122 functions were formatted into the description of functions in the model of Lample et al. \cite{Lample2020Deep} (string) and the subtree proposed in this study.

 The 12,122 functions were divided into strings of tokens, i.e., the smallest units that make sense as mathematical symbols (Supplementary Table 10).
 We created a string polish dataset in which the functions were expressed as strings in Polish notation and a string IRPP dataset in which the integrand was expressed as a string in reverse Polish notation and the primitive function in Polish notation (Supplementary Figure 3).

 To create the subtree data, we transformed the 12,122 functions into an Abstract Syntax Tree (AST) consisting of binary trees using the $parseFormula$ function of libSBML \cite{bornstein2008libsbml}.
 We defined a group of three nodes of the AST as a subtree: a parent node, a left child node, and a right child node.
  The three tokens of the operator or its subject in the order parent node, left child node, and right child node in the subtree were defined as one unit of input to the learning model in the subtree method.
 We then created a subtree polish dataset consisting of strings obtained by a forward search of the ASTs of the integrands and primitive functions per subtree units.
 We also created a subtree IRPP dataset consisting of strings obtained by backward and forward searches of the ASTs of the integrands and primitive functions per subtree units, respectively.
 In this case, if one or both of the left or right child nodes were missing when a node was targeted as the parent node, an ``End of Sentence'' (EOS) token was added to the missing part of the string (Supplementary Figure 3).
\subsection{Learning model}
 The LSTM and Transformer models that we constructed took the integrand as input and output the corresponding primitive function.
 The LSTM model consisted of an embedding layer, an LSTM layer, an attention layer, and a fully connected layer; it was identical to the model in a previous study \cite{Bahdanau2015NeuralMT}, except for the embedding layer and the fully connected layer when the input-output scheme was adopted in a subtree.
 In the subtree method of input, the tokens in the embedding layer represented by one-hot vectors were converted into distributed representations, and then the distributed representations of the three tokens constituting the subtree were concatenated as input to the LSTM layer.
 In the fully connected layer, the output was divided into three parts, each of which was passed to the output function (i.e., softmax) to obtain the corresponding tokens (Supplementary Figure 4).
 Four LSTM models were developed: a model with string polish datasets as input and output (LSTM string polish model), a model with a string IRPP dataset (LSTM string IRPP model), a model with a subtree polish dataset (LSTM subtree polish model), and a model with a subtree IRPP dataset (LSTM subtree IRPP model).
 All hyperparameters in these models were determined by Optuna \cite{akiba2019optuna} with a Tree-structured Parzen Estimator (TPE), which uses Bayesian optimization (Supplementary Table 11).

The Transformer model was composed of an embedding layer: a multi-head attention layer composed of parallel, self-attention layers; a block composed of feed-forward layers; and fully connected layers.
 Except for the embedding layer and the fully connected layer, the rest of the model was the same as the model in a previous study \cite{Lample2020Deep} (Supplementary Figure 5).
 The output of the embedding layer and the fully connected layer in the subtree input were exactly the same as in the LSTM model.
 Four models were developed for the Transformer model and the LSTM model: a model with string polish datasets as input and output (Transformer string polish model), a model with string IRPP datasets (Transformer string IRPP model), a model with subtree polish datasets (Transformer subtree polish model), and a model with subtree IRPP datasets (Transformer subtree IRPP model).
 The hyperparameters of these learning models were identical to the Transformer model of the previous study \cite{Lample2020Deep}, except that the batch size was set to 256 when subtrees were used as the input-output scheme.

 We could assess the correctness of the output of these eight models by differentiating the output of each model using the $D$ function of Mathematica and checking whether the result was equivalent to the integrand.
 We used this strategy to build a model, Integrated All models, that selected the correct results from the output of all eight models.
\subsection{Training and evaluation of the developed learning model}
 All the models developed in this study were trained using 10-fold cross-validation.
 The 12,122 functions, which were pairs of integrands and primitive functions, were divided into two groups of 9,697 and 2,425 functions (4:1 ratio).
 The former was used as the training data, and the latter as the test data.
 The training data were further split at 9:1.
 The latter dataset contained the validation data.
 The training of each learning model was performed with the training data (except for the validation data) based on the Softmax Cross-Entropy error function represented by the following equation:
\begin{eqnarray}
 L = - \frac{1}{m}\sum_{ij}t_{ij} \log y_{ij} \label{eq:softmax_cross_entropy_loss} \\
 y_{ij}=\frac{\exp(x_{ij})}{\sum^n_{k=1}\exp(x_{ik})} \nonumber 
\end{eqnarray}
 In equation (\ref{eq:softmax_cross_entropy_loss}), the input to the activation function softmax was ${\bf X}\in {\bf R}^{m \times n}$, the output was ${\bf Y}\in {\bf R}^{m \times n}$, and the training data were ${\bf T}\in {\bf R}^{m \times n}$.
 Here, $m$ is the number of minibatches and $n$ is the number of dimensions of the hidden state vector.

 The training was carried out for 200 epochs for the LSTM model, 600 epochs for the Transformer string model, and 300 epochs for the Transformer subtree model.
 We adopted the training model that most accurately integrated the validation data.
 The metric of accuracy was the percentage of correct answers, which was 100 times the number of functions obtained by differentiating the output of the training model with the $D$ function in Mathematica that agreed with the input function divided by the total number of validated data.
 This process was carried out until all the training data had been validated. The training model that most accurately integrated the validation data was adopted as the trained model.

The learning model was evaluated by calculating the fraction of correct answers to the test data using the trained model.
\subsection{Evaluation of existing symbolic integration tools}
 To verify the superiority of the developed learning model for symbolic integration, we performed symbolic integrations using existing non-learning-based tools such as Mathematica, Maxima, Rubi, and Matlab.
 We evaluated all the tools by simplifying the integrands for the test data, differentiating the output primitive function with the $D$ function in Mathematica, assessing the agreement with the input integrands, and calculating the fraction of correct answers.
 For Mathematica's symbolic integration, the correct answer rate was calculated with a primitive function obtained by using the $Integrate$ function with the integrand as input, and its derivative was simplified with the $Simplify$ function.
 If no output was obtained after 12 hours, we judged that symbolic integration had failed.
 For Maxima, the correct answer rate was calculated using the primitive function obtained by using the $integrate$ function with the integrand as input, and its derivative was simplified with the $ratsimp$ function.
 For Rubi, the correct answer rate was calculated using the primitive function obtained by using the $Int$ function with the integrand as input, and its derivative was simplified with the $Simplify$ function.
 For Matlab, the correct answer rate was calculated using the primitive function obtained by using the $int$ function with the integrand as input, and the $simplify$ function was used to simplify its derivative.
 The superiority of the developed learning model was verified by comparing the correct answer rates of the symbolic integration of the test data obtained with the learning model to those obtained with non-learning-based tools.
\subsection{Analysis of learning mechanisms using LSTM and Transformer models}
 We found that the symbolic integration problems that were correctly answered differed between the eight models developed in this study, and especially between the LSTM and Transformer models.
 These learning models interpreted the given integrand and transformed it into a primitive function based on different learning mechanisms.
 To better understand the cause of this difference, we analyzed the attention layer in each model to clarify on which part of the integrand the symbolic integration focused.

 For the LSTM model, we visualized the attention layer, which contained information about which part of a given integrand was targeted when the LSTM model was trained on the string polish and subtree polish datasets.
 For each input or output unit of the integrand and primitive function to the LSTM model (per token for the string polish dataset and per subtree for the subtree polish dataset), we visualized the weights of the LSTM model that connected them.
 By visualizing these weights, we could determine which tokens or subtrees of the integrand were targeted when the tokens or subtrees contained in a particular primitive function were output.

 For the Transformer model, we visualized the self-attention layer of the encoder from two perspectives: 1) whether the attention heads within the layer containing the attention head and between the layers related to sequential updating had similar roles, and 2) whether the attention map of the head was uniform from layer to layer.

 To clarify whether there was a similar division of roles among the attention heads within the layer containing the attention head and among the layers related to sequential updating, we mapped the attention map of each attention head into a two-dimensional space using multidimensional scaling \cite{kruskal1964nonmetric} with JS divergence as the dissimilarity.
 The calculation of self-attention was represented by the following equation as in the Transformer string model of a previous study \cite{vaswani2017attention}:
\begin{eqnarray}
 {\rm Attention}({\bf Q,K,V})={\rm softmax}\left(\frac{\bf QK^T}{\sqrt{d_k}}\right )\bf V
\label{self-attention_eq}
\end{eqnarray}
 The symbols ${\bf Q}, {\bf K}$, and ${\bf V}$ represent the matrix of the query vector of each token or subtree of the input functions, the matrix of the key vector, and the matrix of the value vector, respectively.
 If  ${\bf Q}, {\bf K}, and {\bf V}$ are the first stage of the encoder, the first stage represents the matrix ${\bf X}$ of distributed representations of the input tokens or subtrees multiplied by the weights ${\bf W^Q}, {\bf W^K}$, and ${\bf W^V}$, respectively (i.e., ${\bf XW^Q}, {\bf XW^K}$, and ${\bf XW^V}$).
 ${\bf K^T}$ means the transpose of matrix {\bf K} and $d_k$ is the dimensionality of the key vector.
 The second and subsequent layers of the encoder were obtained by multiplying the output of the previous layer by another weight, ${\bf W^Q}, {\bf W^K}$, and ${\bf W^V}$, specific to that layer.
 The dimensions of ${\bf Q}$ and ${\bf K}$ are the numbers of series tokens or subtrees multiplied by $d_k$.
 The dimension of ${\bf V}$ is the number of series tokens or subtrees multiplied by $d_v$, the number of dimensions of the value vector.
 By making a heat map of the weight values, which were the output of the softmax function expressed by equation (\ref{self-attention_eq}) in self-attention, we could verify for a given head which token or subtree of the given integrand in self-attention was targeted and how the targets were sequentially updated.
 In this study as well as previously, we used Multi-head attention \cite{vaswani2017attention}, a method that computes the value from equation (\ref{self-attention_eq}) in multiple parallels to learn more diverse expressions, as follows:
\begin{eqnarray}
 {\rm MultiHead}({\bf Q,K,V}) = {\rm concat}({\rm head_1,\cdots,head_h}){\bf W^O} \label{Multi_head_attention_eq} \\ 
 where\ \ {\rm head}_i = {\rm Attention}\left({\bf QW_i^Q, KW_i^K, VW_i^V}\right ) \nonumber
\end{eqnarray}
 In equation (\ref{Multi_head_attention_eq}), the dimensionalities of the weights are ${\bf W_i^Q} \in {\bf R}^{d_{\rm model}\times d_k}$, ${\bf W_i^K} \in {\bf R}^{d_{\rm model}\times d_k}$, ${\bf W_i^V} \in {\bf R}^{d_{\rm model}\times d_v}$, and ${\bf W^O}\in{\bf R}^{ d_v\times d_{\rm model}}$, where $h$ is the number of attention heads, and $d_{\rm model}$ is the  maximum series length that the model could receive.
 In this way, we tested whether the coordinates on the two-dimensional space of each attention-head in the self-attention layer were close together (i.e., similar in their role assignment) or far apart (i.e., different in their role assignment).

 We then calculated the average entropy of the attention map of the self-attention layer proposed previously to test whether the attention map of the head was homogeneous across layers \cite{clark2019does}.
 In this study, we calculated the entropy of the attention map for each token or subtree in the attention head, and the average entropy of the attention map of the head in the same layer was calculated as the mean entropy.
 The increase in the average entropy through the layers was used to determine whether the attention map was homogenous through the layers.
\subsection{Measurement of the runtime of each learning model and existing non-learning based symbolic integration tools}
 To verify that the symbolic integration based on the learning models developed in this study did not result in unrealistic execution times, we checked the average execution time for a single symbolic integration of the test data using the eight learning models and existing symbolic integration tools.
 Symbolic integration in the learning model was performed on an NVIDIA Tesla V100.
 Symbolic integration in existing non-learning-based tools was performed on a macOS High Sierra 2.2 GHz Intel Core i7.
\section*{Acknowledgement}
 The research was funded by a Japan Science and Technology Agency CREST grant (Grant Number: JPMJCR2011) to A.F.
 Computations were primarily performed using the computing facilities at the University of Tokyo (Reedbush).
 We are grateful for editing the manuscript carefully by two native-English-speaking professional editors from ELSS, Inc.
\section*{Author contributions}
 A.F. planned, managed, and coordinated the entire project as well as supervised the research with T.G.Y.
 H.K. created the dataset with assistance from Y.T., T.G.Y, and A.F.
 H.K. formalized all methods and implementation advised by Y.T. and A.F, and H.K. performed evaluations advised by Y.T. and T.G.Y.
 H.K. validated and visualized all results with advice from Y.T. and T.G.Y.
 H.K. wrote the original manuscript with suggestions from the other authors. The manuscript was translated into English by T.G.Y.
 All authors read and approved the manuscript.
\section*{Competing Interests}
The authors declare no competing interests.

\bibliography{sci}
\bibliographystyle{naturemag}


\end{document}


\maketitle
\begin{affiliations}
 \item Center for Biosciences and Informatics, Graduate School of Fundamental Science and Technology, Keio University, Kanagawa, Japan.
 \item Department of Biosciences and Informatics, Keio University, Yokohama, Kanagawa, Japan.
 \item[*] funa@bio.keio.ac.jp
\end{affiliations}
\listoffigures
\listoftables
\clearpage
\section*{Supplementary Figures}
\begin{figure}[ht]
\begin{center} 
 \includegraphics[width=15cm]{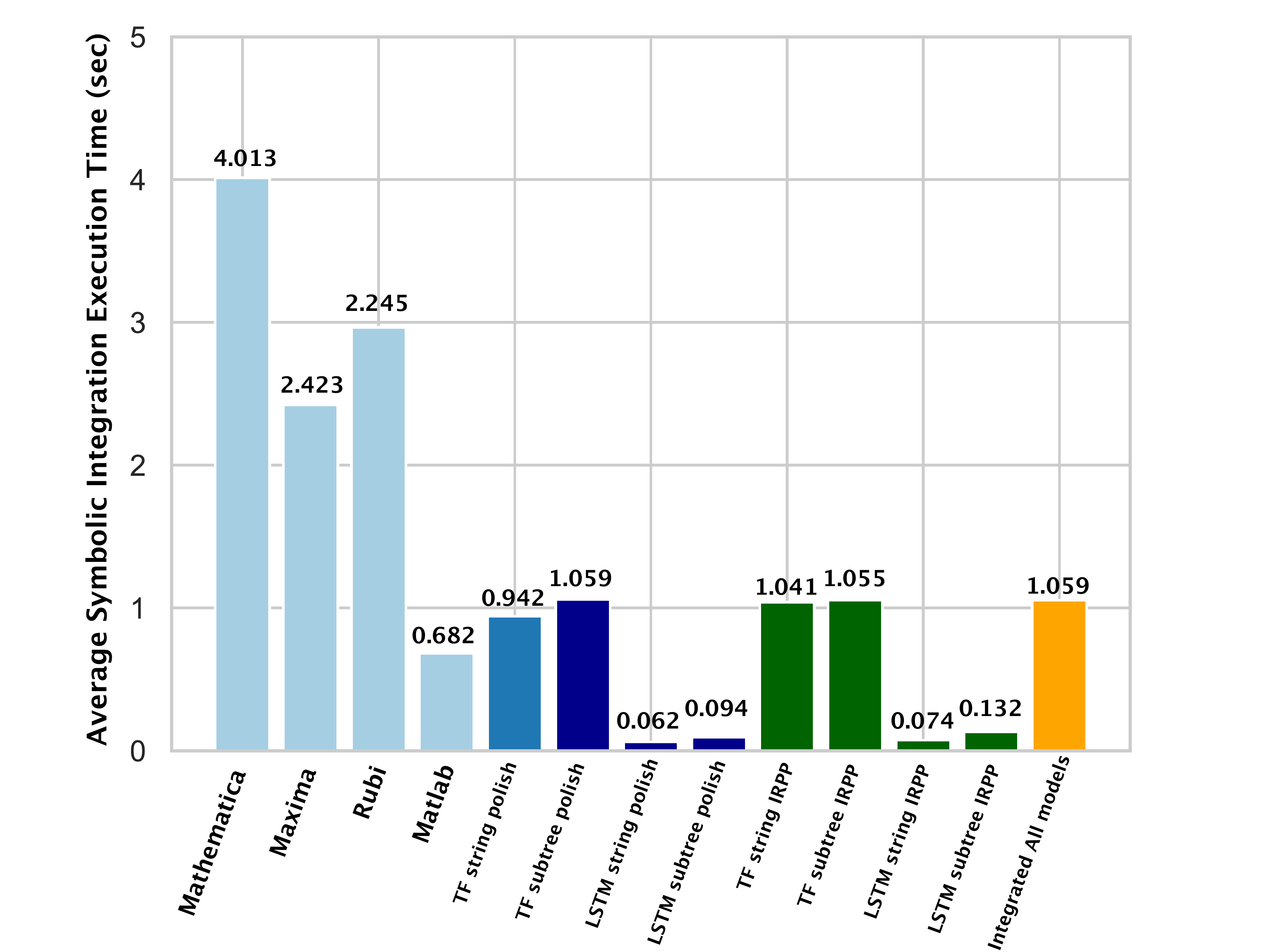}
 \caption[Average runtime for symbolc integration of existing non-learning
 based tools and learning models.]{Average runtime for symbolc integration of existing non-learning
 based tools and learning models. The execution time for symbolic integration
 of the test data using existing non-learning-based tools (light blue, Mathematica,
 Maxima, Rubi, Matlab), the deep learning models developed by Lample et al.
 \cite{Lample2020Deep} (blue, TF string polish as Transformer string polish),
 the learning models developed in this study (dark blue and green), and the
 Integrated All models (orange) are shown as the average run time required for
 the mathematical integration of the test data.}
 \label{fig:timeconsume}
\end{center}
\end{figure}

\begin{figure}[ht]
  \begin{center}
   \includegraphics[width=15cm]{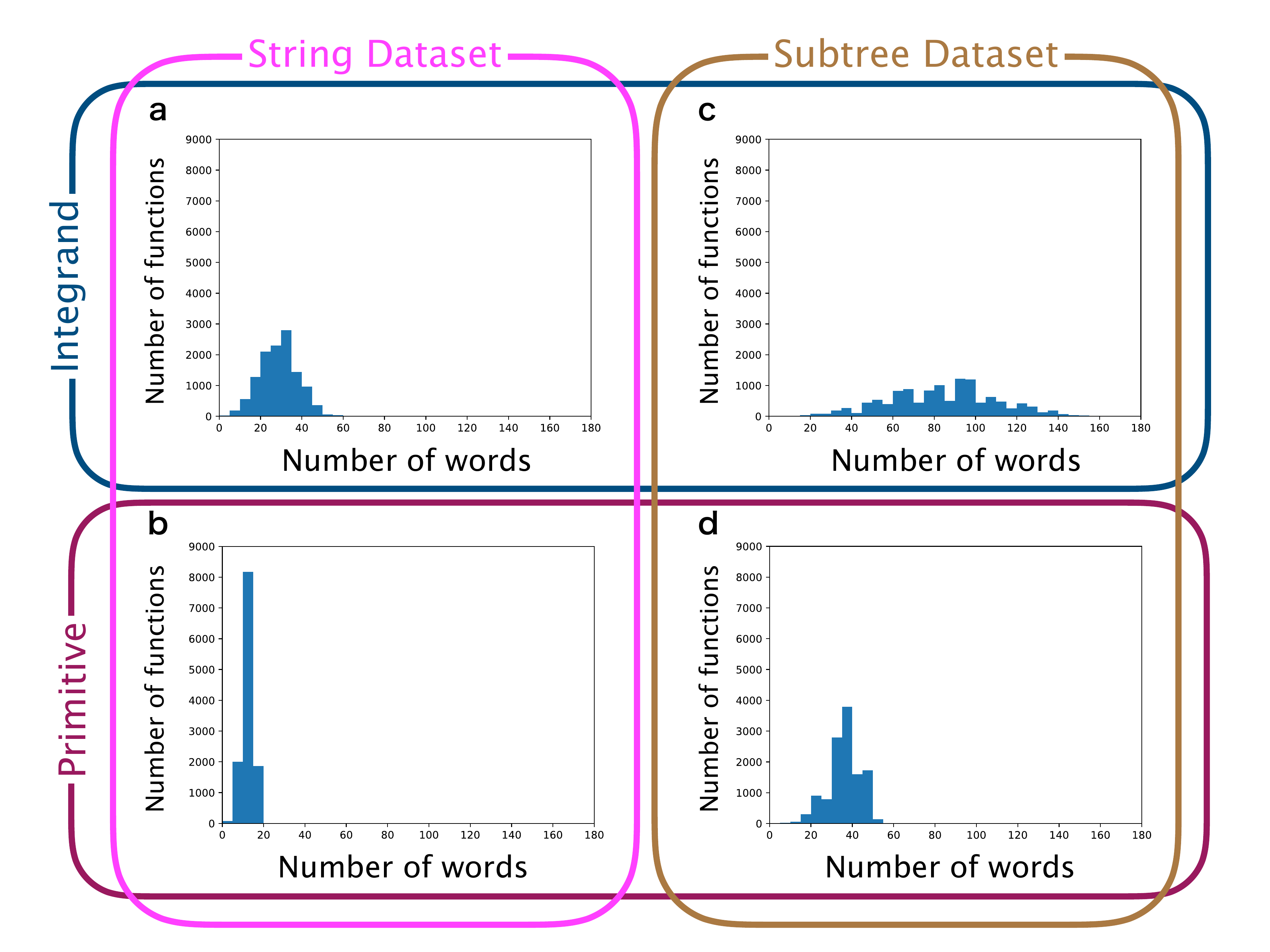}
  \caption[Length distribution of a newly created integrands and primitive functions.]{Length distribution of a newly created integrands and primitive functions.
   Histograms of the number of tokens or subtrees comprising each newly created
   integrand or primitive function are shown {\bf a}: for the integrands in the
   string dataset, {\bf b}: for the primitive functions in the string dataset,
   {\bf c}: for the integrands in the subtree dataset, and {\bf d}: for the
   primitive functions in the subtree dataset.}
  \label{fig:dataset_length}
 \end{center}
\end{figure}

\begin{figure}[ht]
 \begin{center}  
  \includegraphics[width=15cm]{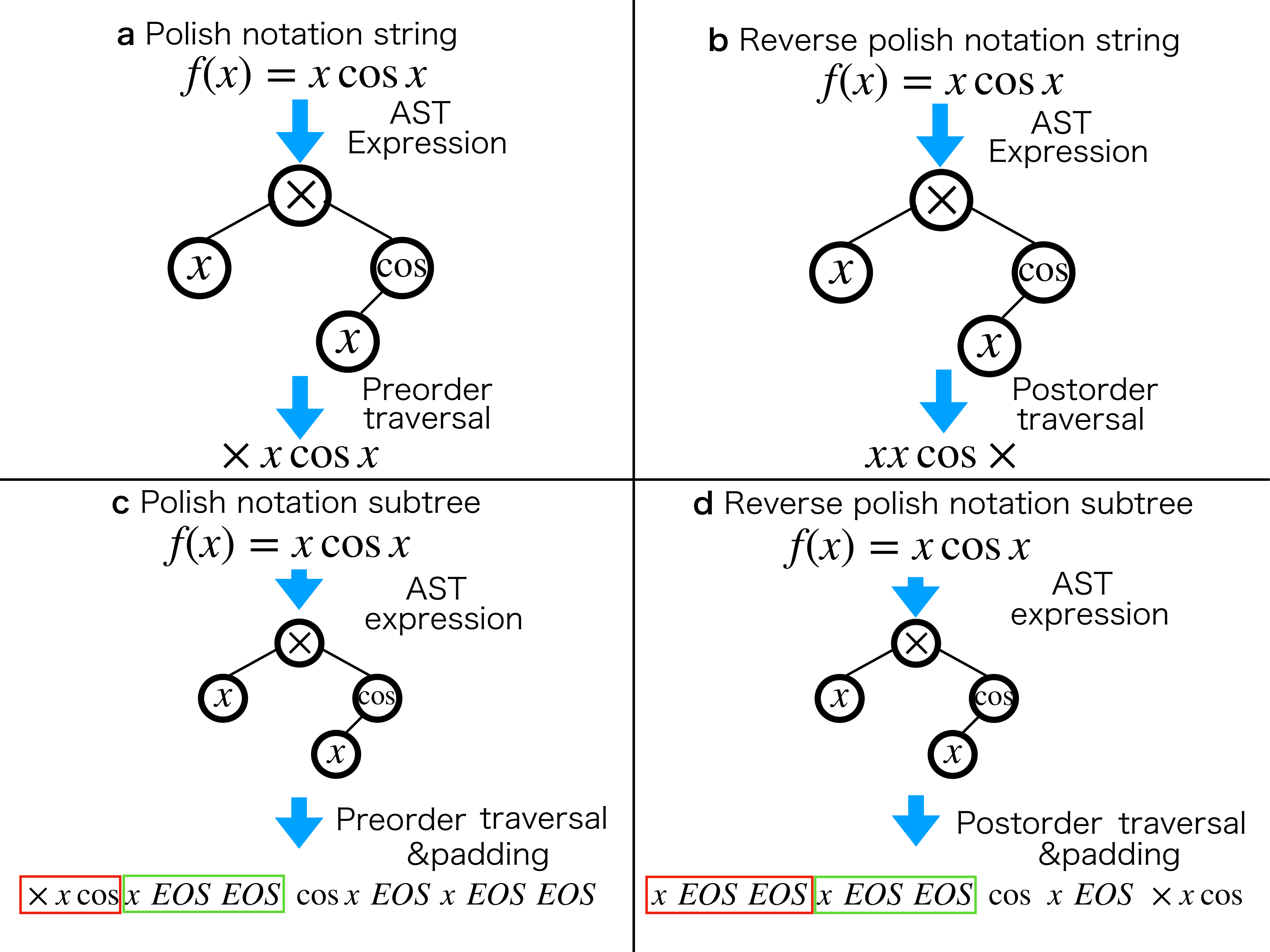}
\caption[Representation based on string/subtree and/or Polish
  notation/reverse Polish notation.]{Representation based on string/subtree and/or Polish
  notation/reverse Polish notation. Conceptual diagram of the description
  of an input mathematical expression ($x \cos x$) from an Abstract Syntax
  Tree (AST) in strings or subtrees in Polish or reverse Polish notation.
  Examples of {\bf a}: writing in Polish notation using the string input
  method (``$\times\ x\ \cos\ x$''), {\bf b}: writing in reverse Polish
  notation using the string input method (``$x\ x\ \cos\ \times$''), {\bf c}:
  writing in Polish notation with the subtree input method
  (``$\times\ x\ \cos\ x\ EOS\ EOS\ \cos\ x\ EOS\ x\ EOS\ EOS$''),
  and {\bf d}: writing in reverse Polish notation with the subtree input
  method (``$x\ EOS\ EOS\ x\ EOS\ EOS\ \cos\ x\ EOS\ \times\ x\ \cos$'').
  In the case of the subtree input method, the string is composed of subtrees
  (i.e., every three tokens in the string input method [red and green boxes]).}
  \label{fig:dataset_description}
 \end{center}
\end{figure}

\begin{figure}[ht]
 \begin{center}
  \includegraphics[width=15cm]{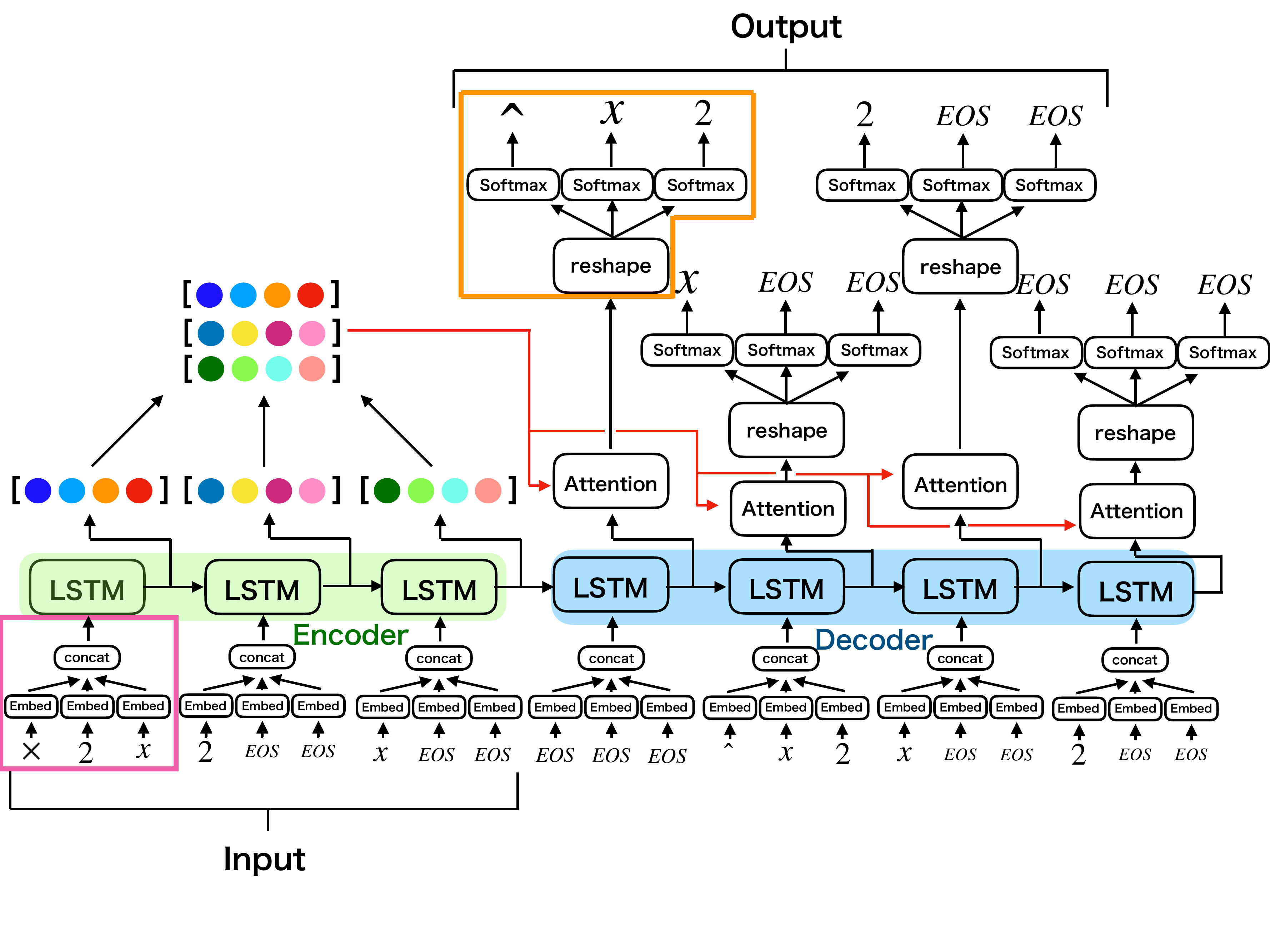}
  \caption[Overview of the Long Short-Term Memory (LSTM) model with
  subtrees as input.]{Overview of the Long Short-Term Memory (LSTM) model with
  subtrees as input. In the embedding layer in the subtree method of input,
  the tokens represented by the one-hot vectors were converted into a
  distributed representation, and then the distributed representations of the
  three tokens constituting the subtree were concatenated as input to the LSTM
  layer (an example is shown in the pink box). In the fully connected layer, the
  output was divided into three parts, each of which was passed to the output
  function (the $softmax$ function) to obtain the corresponding token
  (an example is shown in the orange box).}
  \label{fig:LSTM_subtree_overview}
 \end{center}
\end{figure}

\begin{figure}[ht]
 \begin{center}
  \includegraphics[width=15cm]{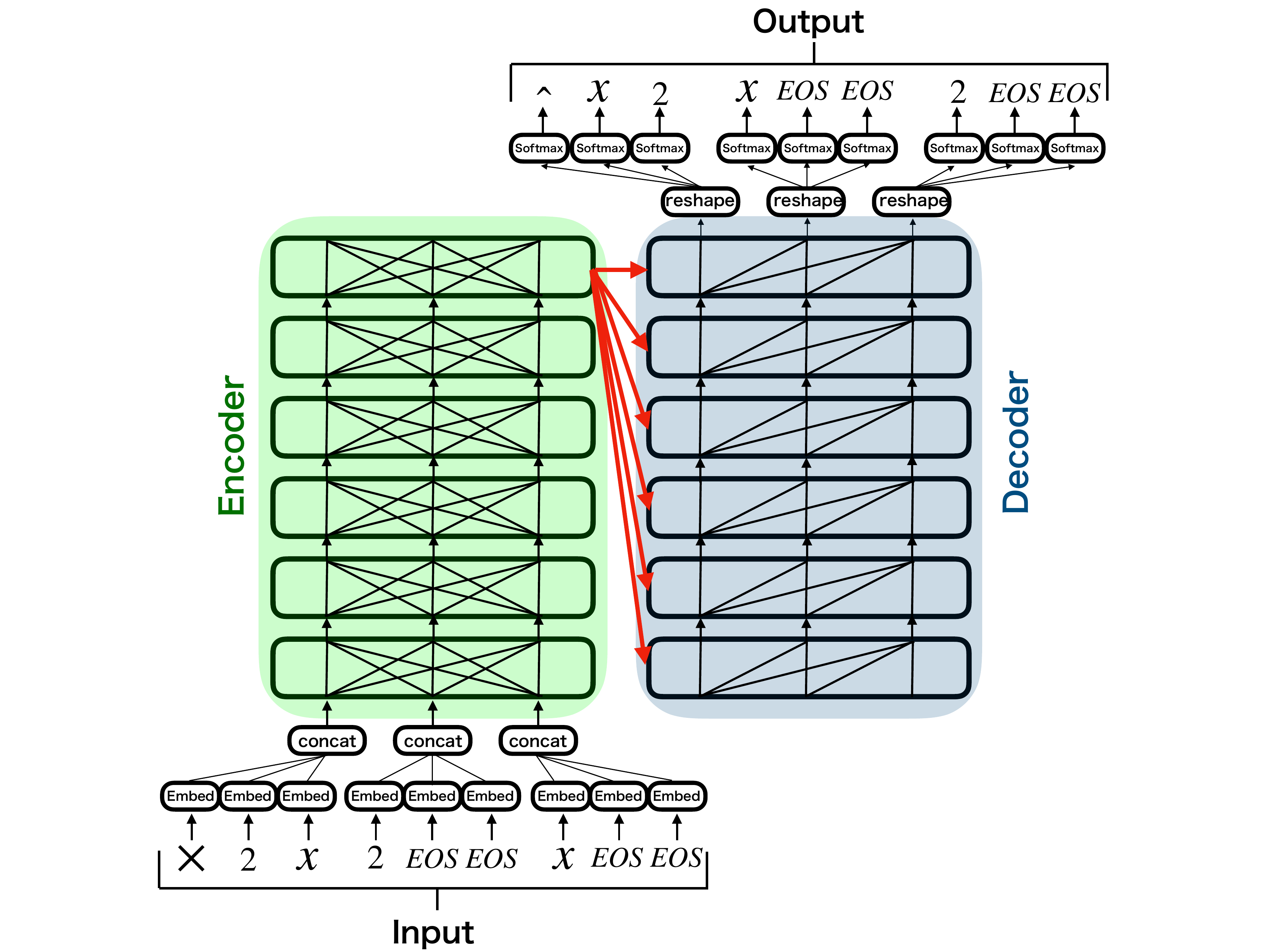}
  \caption[Overview of the Transformer model with subtrees as input.]{Overview of the Transformer model with subtrees as input.
  As in the case of the embedding and fully connected layers in the Long
  Short-Term Memory model with subtree as input, the embedding layer at
  the input converted the tokens represented by the one-hot vector into a
  distributed representation; it then concatenated the distributed
  representations of the three tokens constituting the subtree and used them
  as input to the self-attention layer of the Encoder part.
  The output from the decoder was divided into three parts, and each part
  was passed to the $softmax$ function, which is the output function, to
  obtain the corresponding token. Other than these differences, the structure
  of the Transformer was identical to that of the previous study \cite{Lample2020Deep}.}
  \label{fig:Transformer_subtree_overview}
 \end{center}
\end{figure}
\clearpage
\section*{Supplementary Tables}
\begin{table}[h]
 \caption{Pairs of integrands and primitive functions that were incorrectly answered by learning model with string or subtree.}
 \begin{center}
  \renewcommand{\arraystretch}{2.5}
  \scalebox{0.8}{
  \begin{tabular}{cc||cc}\hline 
  \multicolumn{2}{c||}{String} & \multicolumn{2}{c}{Subtree} \\ \hline 
  Integrands & Primitive functions & Integrands & Primitive funtions \\ \hline
   $-\frac {\cot x \csc x \log x (\cos x (-2+\log x) +2x\csc x \log x )}{x^2}$ & $\frac{\cot^2 x \log^2 x }{x}$ & $-\csc^2 x$ & $\cot x$ \\ \hline 
   \multicolumn{2}{c||}{}& $\frac{-7}{2x^{\frac{9}{2}}}$ & $x^{-\frac{7}{2}}$ \\ \cline{3-4}  
   \multicolumn{2}{c||}{}& $n^{-4}$ & $\frac{x}{n^4}$ \\ \cline{3-4}
   \multicolumn{2}{c||}{}& $\frac{n\sqrt{x}\sec{x}(2x(\log x) \sec x+(-2+3\log x )\sin x)}{2\log^2 x}$ & $\frac{nx^{\frac{3}{2}}\tan x }{\log x}$ \\ \cline{3-4} 
   \multicolumn{2}{c||}{}& $-(2+\cos 2x)\csc^4 x$ & $\cot x \csc^2 x$ \\ \cline{3-4}
   \multicolumn{2}{c||}{}& $\frac{-3}{2x^{\frac{5}{2}}}$ & $x^{\frac{-3}{2}}$ \\ \cline{3-4}
   \multicolumn{2}{c||}{}& $\frac{1}{3\sqrt[3]{x}^2}$ & $\sqrt[3]{x}$ \\ \cline{3-4} 
   \multicolumn{2}{c||}{}& $\frac{e^x\cot^3 x\csc x (-7+\cos (2x)+\sin(2x))}{2}$ & $e^x\cos x \cot^3 x$ \\ \cline{3-4} 
   \multicolumn{2}{c||}{}& $n$ & $nx$ \\ \cline{3-4}
   \multicolumn{2}{c||}{}& $\frac{-3}{x^4}$ & $x^{-3}$ \\ \cline{3-4}
   \multicolumn{2}{c||}{}& $\frac{(\sec^2x)(9x+\sin(2x))(\tan^2(x))}{3\sqrt[3]{x}}$ & $\sqrt[3]{x}^2\tan^3 x$ \\ \cline{3-4}
   \multicolumn{2}{c||}{}& $n^3$ & $n^3x$ \\ \cline{3-4}
   \multicolumn{2}{c||}{}& $\frac{n^2(\sec{x})(3+\cos (2x)-\sin(2x))\tan x )}{2e^x}$ & $\frac{n^2(\sin x)(\tan x)}{e^x}$ \\ \hline 
  \end{tabular}
  }
 \end{center}
\end{table}
\clearpage
{\renewcommand{\arraystretch}{2.5}
\begin{table}[b]
 \caption{Output results for each model for functions commonly answered incorrectly by learning model with string.}
 \centering
 {\footnotesize
 \begin{tabular}{ll}
 \hline 
   Integrand & $-\frac {\cot x \csc x \log x (\cos x (-2+\log x) +2x\csc x \log x )}{x^2}$\\ \hline  
   Primitive function& $\frac{\cot^2 x \log^2 x }{x}$\\ \hline 
   Correct Answer (string)& divide\ times\ power\ cot\ x\ 2\ power\ ln\ x\ 2\ x \\ \hline
   LSTM string polish output & divide\ times\ cos\ x\ power\ ln\ x\ 2\ x \\ \hline
   LSTM string IRPP output & divide\ times\ times\ cos\ x\ power\ cot\ x\ 2\ ln\ x\ x \\ \hline 
   Transformer string polish output & divide\ times\ times\ cos\ x\ power\ cot\ x\ 2\ ln\ x\ x \\ \hline 
   Transformer string IRPP output & divide times power cot x 2 csc x x \\ \hline 
 \end{tabular}
 }
\end{table}
}
\clearpage
{\renewcommand{\arraystretch}{2.5}
{\footnotesize
\begin{longtable}{ll}
 \caption{Output results for each model for functions commonly answered incorrectly by learning model with subtree.}
 \\
 \hline
   Integrand & $-\csc^2 x$ \\ \hline 
   Primitive function& $\cot x$ \\ \hline
   Correct Answer (subtree)& cot\ x\ EOS\ x\ EOS\ EOS \\ \hline
   LSTM subtree polish output& csc\ x\ EOS\ x\ EOS\ EOS \\ \hline
   LSTM subtree IRPP output& csc\ x\ csc\ x\ EOS\ EOS\ \\ \hline  
   Transformer subtree polish output & ln\ x\ eos\ x\ EOS\ EOS \\ \hline
   Transformer subtree IRPP output & csc csc csc csc x EOS x EOS EOS \\ \hline \hline
   Integrand & $\frac{-7}{2x^{\frac{9}{2}}}$ \\ \hline
   Primitive function & $x^{-\frac{7}{2}}$ \\ \hline
   Correct Answer (subtree)& power\ x\ divide\ x\ EOS\ EOS\ divide\ -7\ 2\ -7\ EOS\ EOS\ 2\ EOS\ EOS \\ \hline   
   LSTM subtree polish output & divide\ 1\ power\ 1\ EOS\ EOS\ power\ x\ divide\ x\ EOS\ EOS\ divide\ 7\ 2\ 7\ EOS\ EOS\ 2\ EOS\ EOS \\ \hline
   LSTM subtree IRPP output & divide\ 1\ -2\ 1\ EOS\ EOS\ -2\ x\ divide\ x\ EOS\ EOS\\ \hline 
   Transformer subtree polish output & power\ x\ divide\ x\ EOS\ EOS\ divide\ 7\ 2\ 7\ EOS\ EOS\ 2\ EOS\ EOS\\ \hline
   Transformer subtree IRPP output & power x divide x EOS EOS divide -5 2 -5 EOS EOS 2 EOS EOS \\ \hline \hline  
   Integrand  & $n^{-4}$ \\ \hline
   Primitive function & $\frac{x}{n^4}$ \\ \hline 
   Correct Answer (subtree)& divide x power x EOS EOS power n 4 n EOS EOS 4 EOS EOS\\ \hline
   LSTM subtree polish output & divide x power x EOS EOS power n 2 n EOS EOS 2 EOS EOS\\ \hline
   LSTM subtree IRPP output & divide x power x EOS EOS power n 3 n EOS EOS 3 EOS EOS \\ \hline
   Transformer subtree polish & divide x power x EOS EOS power n 3 n EOS EOS 3 EOS EOS \\ \hline
   Transformer subtree IRPP & divide x power x EOS EOS power n 3 n EOS EOS 3 EOS EOS \\ \hline \hline 
   \pagebreak 
   \hline
   Integrand & $\frac{n\sqrt{x}\sec{x}(2x(\log x) \sec x+(-2+3\log x )\sin x)}{2\log^2 x}$ \\ \hline
   Primitive function & $\frac{nx^{\frac{3}{2}}\tan x }{\log x}$ \\ \hline
   Correct Answer (subtree) & divide times ln times times tan times n power n EOS EOS power x divide x EOS EOS \\ 
                            & divide 3 2 3 EOS EOS 2 EOS EOS tan x EOS x EOS EOS ln x EOS x EOS EOS \\ \hline
   LSTM subtree polish output & divide times ln times times tan times times sec times n power n EOS EOS power x divide \\ 
                              & x EOS EOS divide 3 2 3 EOS EOS 2 EOS EOS sec x EOS x EOS EOS \\ 
                              & tan x EOS x EOS EOS ln x EOS x EOS EOS \\ \hline
   LSTM subtree IRPP output & divide times ln times times sec times n power n EOS EOS power x divide x EOS EOS  \\
                              & divide 3 2 3 EOS EOS 2 EOS EOS sec x EOS x EOS EOS ln x EOS x EOS EOS \\ \hline
   Transformer subtree polish output & divide times ln times times sec times n power n EOS EOS power x divide x EOS EOS \\
                                     & divide 3 2 3 EOS EOS 2 EOS EOS sec x EOS x EOS EOS ln x EOS x EOS EOS \\ \hline
   Transformer subtree IRPP output & divide times ln times times tan times n sin n EOS EOS sin x EOS x EOS EOS \\
                                   & tan x EOS x EOS EOS ln x EOS x EOS EOS \\ \hline  
   \pagebreak 
   \hline 
   Integrand & $-(2+\cos 2x)\csc^4 x$ \\ \hline
   Primitive function & $\cot x \csc^2 x$ \\ \hline
   Correct Answer (subtree) & times cot power cot x EOS x EOS EOS power csc 2 csc x EOS x EOS EOS 2 EOS EOS \\ \hline
   LSTM subtree polish output & times power power power cos 2 cos x EOS x EOS EOS 2 EOS EOS power csc 2 \\ 
                              & csc x EOS x EOS EOS 2 EOS EOS \\ \hline 
   LSTM subtree IRPP output & times power power power cot 2 cot x EOS x EOS EOS 2 EOS EOS power csc 2 \\ 
                            & csc x EOS x EOS EOS 2 EOS EOS \\ \hline
   Transformer subtree polish output & times power power power cot 2 cot x EOS x EOS EOS 2 EOS EOS power csc 2 \\ 
                            & csc x EOS x EOS EOS 2 EOS EOS \\ \hline
   Transformer subtree IRPP output & times power power power cot 2 cot x EOS x EOS EOS 2 EOS EOS power csc 2 \\ 
                            & csc x EOS x EOS EOS 2 EOS EOS \\ \hline \hline 
   Integrand & $\frac{-3}{2x^{\frac{5}{2}}}$ \\ \hline
   Primitive function & $x^{\frac{-3}{2}}$ \\ \hline 
   Correct Answer (subtree) &  power x divide x EOS EOS divide -3 2 -3 EOS EOS 2 EOS EOS \\ \hline
   LSTM subtree polish output & divide 1 power 1 EOS EOS power x divide x EOS EOS divide 3 2 3 EOS EOS 2 EOS EOS \\ \hline
   LSTM subtree IRPP output & divide 1 divide 1 EOS EOS -2 x divide x EOS EOS divide 7 2 7 EOS EOS 2 EOS EOS \\ \hline
   Transformer subtree polish output & power 1 divide x EOS EOS divide 3 2 3 EOS EOS 2 EOS EOS \\ \hline
   Transformer subtree IRPP output & power x divide x EOS EOS divide -5 2 7 EOS EOS 2 EOS EOS \\ \hline 
   \pagebreak 
   \hline 
   Integrand & $\frac{1}{3\sqrt[3]{x}^2}$ \\ \hline
   Primitive function & $\sqrt[3]{x}$ \\ \hline
   Correct Answer (subtree) & root 3 x 3 EOS EOS x EOS EOS \\ \hline
   LSTM subtree polish output & power root root root 3 x 3 EOS EOS x EOS EOS -3 EOS EOS \\ \hline
   LSTM subtree IRPP output & power root root root 3 x 3 EOS EOS x EOS EOS x EOS EOS \\ \hline
   Transformer subtree polish output & power root root root 3 x 3 EOS EOS x EOS EOS 2 EOS x \\ \hline
   Transformer subtree IRPP output & power root root root 3 x 3 EOS EOS x EOS EOS -1 EOS EOS \\ \hline \hline  
   Integrand & $\frac{e^x\cot^3 x\csc x (-7+\cos (2x)+\sin(2x))}{2}$ \\ \hline 
   Primitive function& $e^x\cos x \cot^3 x$ \\ \hline
   Correct Answer (subtree) & times times power times power cos power e x e EOS EOS x EOS EOS cos x EOS \\ 
                            & x EOS EOS power cot 3 cot x EOS x EOS EOS 3 EOS EOS \\ \hline
   LSTM subtree polish output & times times power times power power power e x e EOS EOS x EOS EOS power cos 2 \\ 
                              & cos x EOS x EOS EOS 2 EOS EOS power cot 2 cot x EOS x EOS EOS 2 EOS EOS \\ \hline
   LSTM subtree IRPP output & times times power times power cos power e x e EOS EOS x EOS EOS cos x EOS \\ 
                            & x EOS EOS power cot 2 cot x EOS x EOS EOS 2 EOS EOS \\ \hline
   Transformer subtree polish output & times times power times power cos power e x e EOS EOS x EOS EOS cos x EOS \\ 
                                     & x EOS EOS power cot 2 cot x EOS x EOS EOS 2 EOS EOS \\ \hline
   Transformer subtree IRPP output & times times power times power cos power e x e EOS EOS x EOS EOS cos x EOS \\
                                   & x EOS EOS power cot 2 cot x EOS x EOS EOS 2 EOS EOS \\ \hline 
   \pagebreak 
   \hline   
   Integrand& $n$ \\ \hline 
   Primitive function & $nx$ \\ \hline
   Correct Answer (subtree) & times n x n EOS EOS x EOS EOS \\ \hline
   LSTM subtree polish output & divide n n n EOS EOS n EOS EOS \\ \hline
   LSTM subtree IRPP output &  n EOS EOS \\ \hline
   Transformer subtree polish output & divide power n power EOS 2 n EOS EOS 2 EOS EOS n EOS EOS \\ \hline
   Transformer subtree IRPP output & n n n n EOS EOS \\ \hline \hline 
   
   Integrand & $\frac{-3}{x^4}$ \\ \hline
   Primitive function & $x^{-3}$ \\ \hline
   Correct Answer (subtree) & power x -3 x EOS EOS -3 EOS EOS \\ \hline
   LSTM subtree polish output & power x 5 x EOS EOS -1 EOS EOS \\ \hline
   LSTM subtree IRPP output & power x 3 x EOS EOS 3 EOS EOS \\ \hline
   Transformer subtree polish output & power x power x EOS EOS power EOS 3 3 x EOS x EOS EOS \\ \hline
   Transformer subtree IRPP output & power ln -3 ln x EOS x EOS EOS -3 EOS EOS \\ \hline 
   \pagebreak
   \hline
   Integrand & $\frac{(\sec^2x)(9x+\sin(2x))(\tan^2(x))}{3\sqrt[3]{x}}$ \\ \hline 
   Primitive function & $\sqrt[3]{x}^2\tan^3 x$ \\ \hline
   Correct Answer (subtree) & times power power power root 2 root 3 x 3 EOS EOS x EOS EOS 2 EOS EOS \\ 
                            & power tan 3 tan x EOS x EOS EOS 3 EOS EOS \\ \hline
   LSTM subtree polish output & times times power times x power x EOS EOS power root 3 root 3 x 3 EOS EOS \\ 
                              & x EOS EOS 3 EOS EOS power tan 2 tan x EOS x EOS EOS 2 EOS EOS \\ \hline
   LSTM subtree IRPP output & times power power power tan 3 tan x EOS x EOS EOS 3 EOS EOS power root 2 \\ 
                            & root 3 x 3 EOS EOS x EOS EOS 2 EOS EOS \\ \hline
   Transformer subtree polish output & times power power power root 3 root 3 x 3 EOS EOS x EOS EOS 3 EOS EOS \\ 
                                     & power tan 3 tan x EOS x EOS EOS 3 EOS EOS \\ \hline
   Transformer subtree IRPP output & times power power power sec 3 sec x EOS x EOS EOS 3 EOS EOS power root 2 \\
                                   & root 3 x 3 EOS EOS x EOS EOS 2 EOS EOS \\ \hline \hline 
   Integrand & $n^3$ \\ \hline
   Primitive function & $n^3x$ \\ \hline
   Correct Answer (subtree) & times power x power n 3 n EOS EOS 3 EOS EOS x EOS EOS \\ \hline
   LSTM subtree polish output & divide power power power n 2 n EOS EOS 2 EOS EOS power n 3 n EOS EOS 3 EOS EOS \\ \hline
   LSTM subtree IRPP output & times power power power n 2 n EOS EOS 2 EOS EOS power x 3 x EOS EOS 3 EOS EOS \\ \hline
   Transformer subtree polish output & times power power power n 3 n EOS EOS 3 EOS EOS power x 2 x EOS EOS 2 EOS EOS \\ \hline
   Transformer subtree IRPP output & times power power power n 3 n EOS EOS 3 EOS EOS power root 2 root 3 x \\
                                   & 3 EOS EOS x EOS EOS 2 EOS EOS \\ \hline 
   \pagebreak
   \hline  
   Integrand & $\frac{n^2(\sec{x})(3+\cos (2x)-\sin(2x))\tan x}{2e^x}$ \\ \hline
   Primitive function & $\frac{n^2(\sin x)(\tan x)}{e^x}$ \\ \hline
   Correct Answer (subtree) & divide times power times times tan times power sin power n 2 n EOS EOS 2 EOS EOS \\ 
                            & sin x EOS x EOS EOS tan x EOS x EOS EOS power e x e EOS EOS x EOS EOS \\ \hline
   LSTM subtree polish output & divide times power times times power times power sec power n 2 n EOS EOS 2 EOS EOS \\ 
                              & sin x EOS x EOS EOS power tan 2 tan x EOS x EOS EOS 2 EOS EOS \\
                              & power e x e EOS EOS x EOS EOS \\ \hline 
   LSTM subtree IRPP output & divide times power times times power times power sin power n 2 n EOS EOS 2 EOS EOS \\ 
                            & sin x EOS x EOS EOS power tan 2 tan x EOS x EOS EOS 2 EOS EOS \\ 
                            & power e x e EOS EOS x EOS EOS \\ \hline
   Transformer subtree polish output & divide times power times times power times power sin power n 2 n EOS EOS 2 EOS EOS \\
                                     & sin x EOS x EOS EOS power tan 2 tan x EOS x EOS EOS 2 EOS EOS \\
                                     & power e x e EOS EOS x EOS EOS \\ \hline
   Transformer subtree IRPP output & divide times power times times power times power sin power n 2 n EOS EOS 2 EOS EOS \\
                                   & sin x EOS x EOS EOS power tan 2 tan x EOS x EOS EOS 2 EOS EOS \\  
                                   & power e x e EOS EOS x EOS EOS \\ \hline 
\end{longtable}
}
}
\clearpage

\begin{table}[b]
 \caption{Pairs of integrands and primitive functions that were incorrectly answered by learning model adopting Long Short-Term Memory (LSTM) or Transformer.}
 \begin{center}
  \renewcommand{\arraystretch}{2.5}
  \scalebox{0.8}{
  \begin{tabular}{cc||cc}\hline 
  \multicolumn{2}{c||}{LSTM} & \multicolumn{2}{c}{Transformer} \\ \hline 
  Integrands & Primitive functions & Integrands & Primitive funtions \\ \hline
  $-(2+\cos (2x))\csc^4x$ & $\cot x \csc ^2x$ & $-\frac{\sqrt{x}(-7+12x\cot{x})\csc^2 x}{6\sqrt[3]{x}}$ & $\frac{x^{\frac{3}{2}}\csc^2 x}{\sqrt[3]{x}}$ \\ \hline
   $-\frac {(2+\cos (2x))\csc^4(x)}{n}$ & $\frac{\cot x \csc^2 x}{n}$ & $\frac{-7}{2x^{\frac{9}{2}}}$ & $x^{-\frac{7}{2}}$ \\ \hline 
   $n$ & $nx$ & $\frac{-3}{2x^{\frac{5}{2}}}$ & $x^{\frac{-3}{2}}$ \\ \hline    
   $\frac{4(1-12x\cot(2x))\csc^2(2x)}{3\sqrt[3]{x}^2}$& $4\csc^2 (2x)\sqrt[3]{x}$ & $-\frac {\cot x \csc x \log x (\cos x (-2+\log x) +2x\csc x \log x )}{x^2}$ & $\frac{\cot^2 x \log^2 x }{x}$ \\ \hline
  $n^3$ & $n^3x$ & $\frac{(\sec^2x)(9x+\sin(2x))(\tan^2(x))}{3\sqrt[3]{x}}$ & $\sqrt[3]{x}^2\tan^3 x$ \\ \hline
  $\frac{n^2(\sec{x})(3+\cos (2x)-\sin(2x))\tan x )}{2e^x}$ & $\frac{n^2(\sin x)(\tan x)}{e^x}$ &\multicolumn{2}{c}{}  \\ \hline
  \end{tabular}
  }
 \end{center}
\end{table}

\clearpage
{\renewcommand{\arraystretch}{2.5}
{\footnotesize
\begin{longtable}{ll}
 \caption{Output results for each model for functions commonly incorrectly answered by learning model adopting Long Short-Term Memory (LSTM).}
 \\
 \hline 
   Integrand& $-(2+\cos (2x))\csc^4x$ \\ \hline 
   Primitive function & $\cot x \csc ^2x$ \\ \hline
   Correct Answer (string) & times cot x power csc x 2 \\ \hline 
   LSTM string polish output & times cot x power csc x 3 \\ \hline
   LSTM string IRPP output & times cot x power cot x 2 \\ \hline
   Correct Answer (subtree) & times cot power cot x EOS x EOS EOS power csc 2 csc x EOS x EOS EOS 2 EOS EOS \\ \hline
   LSTM subtree polish output & times power power power cos 2 cos x EOS x EOS EOS 2 EOS EOS power csc 2 \\  
                              & csc x EOS x EOS EOS 2 EOS EOS \\ \hline
   LSTM subtree IRPP output & times power power power cot 2 cot x EOS x EOS EOS 2 EOS EOS power csc 2 \\ 
                            & csc x EOS x EOS EOS 2 EOS EOS\\ \hline \hline 

   Integrand & $-\frac {(2+\cos (2x))\csc^4(x)}{n}$ \\ \hline
   Primitive function & $\frac{\cot x \csc^2 x}{n}$ \\ \hline
   Correct Answer (string) & divide times cot x power csc x 2 n \\ \hline
   LSTM string polish output & divide times cot x csc x n \\ \hline
   LSTM string IRPP output & divide times cot x csc x n \\ \hline
   Correct Answer (subtree) & divide times n times cot power cot x EOS x EOS EOS power csc 2 csc x EOS \\ 
                            & x EOS EOS 2 EOS EOS n EOS EOS \\ \hline
   LSTM subtree polish output & divide times n times power power power cos 2 cos x EOS x EOS EOS 2 EOS EOS \\ 
                              & power csc 2 csc x EOS x EOS EOS 2 EOS EOS n EOS EOS \\ \hline
   LSTM subtree IRPP output & divide times n times power power power cot 2 cot x EOS x EOS EOS 2 EOS EOS \\ 
                            & power csc 2 csc x EOS x EOS EOS 2 EOS EOS n EOS EOS \\ \hline 
   \pagebreak 
   \hline 
   Integrand & $n$\\ \hline
   Primitive function & $nx$ \\ \hline
   Correct Answer (string) & times n x \\ \hline
   LSTM string polish output & divide n \\ \hline
   LSTM string IRPP output & EOS \\ \hline
   Correct Answer (subtree) & times n x n EOS EOS x EOS EOS \\ \hline
   LSTM subtree polish output & divide n n n EOS EOS n EOS EOS \\ \hline
   LSTM subtree IRPP output & n EOS EOS \\ \hline \hline 
   Integrand & $\frac{4(1-12x\cot(2x))\csc^2(2x)}{3\sqrt[3]{x}^2}$ \\ \hline
   Primitive function & $4\csc^2 (2x)\sqrt[3]{x}$ \\ \hline
   Correct Answer (string) & times times 4 power csc times 2 x 2 root 3 x \\ \hline
   LSTM string polish output & times times 4 power csc x 2 root 3 x \\ \hline
   LSTM string IRPP output & times times 4 power csc x 2 root 3 x \\ \hline
   Correct Answer (subtree) & times times root times 4 power 4 EOS EOS power csc 2 csc times EOS times 2 x \\ 
                            & 2 EOS EOS x EOS EOS 2 EOS EOS root 3 x 3 EOS EOS x EOS EOS \\ \hline
   LSTM subtree polish output & times times root times csc power csc x EOS x EOS EOS power csc 2 csc x EOS \\ 
                              & x EOS EOS 2 EOS EOS root 3 x 3 EOS EOS x EOS EOS \\ \hline
   LSTM subtree IRPP output & times times root times csc power csc x EOS x EOS EOS power csc 2 csc x EOS \\
                            & x EOS EOS 2 EOS EOS root 3 x 3 EOS EOS x EOS EOS \\ \hline 
   \pagebreak  
   \hline 
   Integrand & $n^3$ \\ \hline
   Primitive function & $n^3x$ \\ \hline
   Correct Answer (string) & times power n 3 x \\ \hline
   LSTM string polish output & divide x power n 3 \\ \hline
   LSTM string IRPP output & divide x power n 3 \\ \hline
   Correct Answer (subtree) & times power x power n 3 n EOS EOS 3 EOS EOS x EOS EOS \\ \hline
   LSTM subtree polish output & divide power power power n 2 n EOS EOS 2 EOS EOS power n 3 n EOS EOS 3 EOS EOS \\ \hline
   LSTM subtree IRPP output & times power power power n 2 n EOS EOS 2 EOS EOS power x 3 x EOS EOS 3 EOS EOS \\ \hline \hline
   Integrand & $\frac{n^2(\sec{x})(3+\cos (2x)-\sin(2x))\tan x }{2e^x}$ \\ \hline
   Primitive function & $\frac{n^2(\sin x)(\tan x)}{e^x}$ \\ \hline
   Correct Answer (string) & divide times times power n 2 sin x tan x power e x \\ \hline
   LSTM string polish output & divide times times power n 2 cos x sin x power e x \\ \hline
   LSTM string IRPP output & divide times times power n 2 sin x power tan x 2 power e x \\ \hline
   Correct Answer (subtree) & divide times power times times tan times power sin power n 2 n EOS EOS 2 EOS EOS \\  
                            & sin x EOS x EOS EOS tan x EOS x EOS EOS power e x e EOS EOS x EOS EOS \\ \hline
   LSTM subtree polish output & divide times power times times power times power sec power n 2 n EOS EOS 2 EOS EOS \\
                              & sin x EOS x EOS EOS power tan 2 tan x EOS x EOS EOS 2 EOS EOS \\ 
                              & power e x e EOS EOS x EOS EOS \\ \hline
   LSTM subtree IRPP output & divide times power times times power times power sin power n 2 n EOS EOS 2 EOS EOS \\ 
                            & sin x EOS x EOS EOS power tan 2 tan x EOS x EOS EOS 2 EOS EOS \\ 
                            & power e x e EOS EOS x EOS EOS \\ \hline
\end{longtable}
}
}
\clearpage
{\renewcommand{\arraystretch}{2.5}
{\footnotesize
\begin{longtable}{ll}
 \caption{Output results for each model for functions commonly incorrectly answered by learning model adopting Transformer.}
 \\
 \hline 
   Integrand & $-\frac{\sqrt{x}(-7+12x\cot{x})\csc^2 x}{6\sqrt[3]{x}}$ \\ \hline 
   Primitive function & $\frac{x^{\frac{3}{2}}\csc^2 x}{\sqrt[3]{x}}$ \\ \hline
   Correct Answer (string) & divide times power x divide 3 2 power csc x 2 root 3 x \\ \hline
   Transformer string polish output & times times root 2 x power csc x 2 power root 3 x 2 \\ \hline
   Transformer string IRPP output & times times root 2 x power csc x 2 power root 3 x 2 \\ \hline
   Correct Answer (subtree) & divide times root times power power power x divide x EOS EOS divide 3 2 3 EOS EOS \\ 
                            & 2 EOS EOS power csc 2 csc x EOS x EOS EOS 2 EOS EOS root 3 x 3 EOS EOS x EOS EOS \\ \hline
   Transformer subtree polish output & times times power times root power root 2 x 2 EOS EOS x EOS EOS power csc 2 \\
                                     & csc x EOS x EOS EOS 2 EOS EOS power root 2 root 3 x 3 EOS EOS x EOS EOS 2 EOS EOS \\ \hline
   Transformer subtree IRPP output & divide times root times times power times root root root 2 x 2 EOS EOS x EOS EOS \\ 
                                   & root 3 x 3 EOS EOS x EOS EOS power csc 2 csc x EOS x EOS EOS \\ 
                                   & 2 EOS EOS root 3 x 3 EOS EOS x EOS EOS \\ \hline \hline 

   Integrand & $\frac{-7}{2x^{\frac{9}{2}}}$ \\ \hline
   Primitive function & $x^{-\frac{7}{2}}$ \\ \hline
   Correct Answer (string) & power x divide minus 7 2 \\ \hline
   Transformer string polish output & power x divide 7 2 \\ \hline
   Transformer string IRPP output & x divide 7 2  \\ \hline
   Correct Answer (subtree) & power x divide x EOS EOS divide -7 2 -7 EOS EOS 2 EOS EOS \\ \hline
   Transformer subtree polish output & power x divide x EOS EOS divide 7 2 7 EOS EOS 2 EOS EOS \\ \hline
   Transformer subtree IRPP output & power x divide x EOS EOS divide -5 2 -5 EOS EOS 2 EOS EOS \\ \hline 
   \pagebreak
   \hline
   Integrand & $\frac{-3}{2x^{\frac{5}{2}}}$ \\ \hline
   Primitive function & $x^{\frac{-3}{2}}$ \\ \hline
   Correct Answer (string)& power x divide minus 3 2 \\ \hline
   Transformer string polish output & power x divide 3 2 \\ \hline
   Transformer string IRPP output & divide 1 power x divide 3 2 \\ \hline
   Correct Answer (subtree) &  power x divide x EOS EOS divide -3 2 -3 EOS EOS 2 EOS EOS \\ \hline
   Transformer subtree polish output & power 1 divide x EOS EOS divide 3 2 3 EOS EOS 2 EOS EOS \\ \hline
   Transformer subtree IRPP output & power x divide x EOS EOS divide -5 2 7 EOS EOS 2 EOS EOS \\ \hline \hline 
   Integrand & $-\frac {\cot x \csc x \log x (\cos x (-2+\log x) +2x\csc x \log x )}{x^2}$ \\ \hline 
   Primitive function & $\frac{\cot^2 x \log^2 x }{x}$ \\ \hline
   Correct Answer (string)& divide times power cot x 2 power ln x 2 x \\ \hline
   Transformer string polish output & divide times power cot x 2 csc x power ln x 2 \\ \hline
   Transformer string IRPP output & divide times power cot x 2 csc x x \\ \hline
   Correct Answer (subtree) & divide times x times power power power cot 2 cot x EOS x EOS EOS 2 EOS EOS \\ 
                            & power ln 2 ln x EOS x EOS EOS 2 EOS EOS x EOS EOS \\ \hline
   Transformer subtree polish output & divide times x times power power power cos 2 cos x EOS x EOS EOS 2 EOS EOS \\
                                     & power ln 2 ln x EOS x EOS EOS 2 EOS EOS x EOS EOS \\ \hline
   Transformer subtree IRPP output & divide times x times cos power cos x EOS x EOS EOS power cot 2 cot x EOS \\
                                   & x EOS EOS 2 EOS EOS x EOS EOS \\ \hline
   \pagebreak 
   \hline  
   Integrand & $\frac{(\sec^2x)(9x+\sin(2x))(\tan^2(x))}{3\sqrt[3]{x}}$ \\ \hline 
   Primitive function & $\sqrt[3]{x}^2\tan^3 x$ \\ \hline 
   Correct Answer (string) & times power root 3 x 2 power tan x 3 \\ \hline
   Transformer string polish output & times power sin x 3 power root 3 x 2 \\ \hline
   Transformer string IRPP output & times power sec x 3 power root 3 x 2 \\ \hline
   Correct Answer (subtree) & times power power power root 2 root 3 x 3 EOS EOS x EOS EOS 2 EOS EOS \\ 
                            & power tan 3 tan x EOS x EOS EOS 3 EOS EOS \\ \hline
   Transformer subtree polish output & times power power power root 3 root 3 x 3 EOS EOS x EOS EOS 3 EOS EOS \\ 
                                     & power tan 3 tan x EOS x EOS EOS 3 EOS EOS \\ \hline
   Transformer subtree IRPP output & times power power power sec 3 sec x EOS x EOS EOS 3 EOS EOS power root 2 \\
                                   & root 3 x 3 EOS EOS x EOS EOS 2 EOS EOS \\ \hline
\end{longtable}
}
}
\clearpage
{\renewcommand{\arraystretch}{2.5}
\begin{table}[b]
 \caption{Pairs of integrands and primitive functions that were incorrectly answered by learning model with polish or Integrand Reverse polish Primitive Polish (IRPP).}
 \begin{center}
  \begin{tabular}{cc||cc}\hline 
  \multicolumn{2}{c||}{Polish} & \multicolumn{2}{c}{IRPP} \\ \hline 
  Integrands & Primitive functions & Integrands & Primitive funtions \\ \hline
   $\frac{-3}{2x^{\frac{5}{2}}}$ & $x^{\frac{-3}{2}}$ & $\frac{\sec x(3x\sec x-2\sin x )\tan^2x}{x^3}$ & $\frac{\tan^3 x}{x^2}$ \\ \hline
   $-\frac{(2+\cos(2x))\csc^4 x}{n}$ & $\frac{\cot x \csc^2x}{n}$ & $n\csc^2 x(-3\csc^2 x + \sec^2 x  )$ & $n\csc^3 x \sec x $ \\ \hline
   \multicolumn{2}{c||}{} & $\frac{1}{3\sqrt[3]{x}^2}$ & $\sqrt[3]{x}$ \\ \hline  
   \multicolumn{2}{c||}{} & $\frac{n^2\sec x (3x+x\cos (2x)-\sin(2x))\tan x}{2x^2}$ & $\frac{n^2\sin x \tan x}{x}$ \\ \hline  
  \multicolumn{2}{c||}{} & $-\frac {\cot x \csc x \log x (\cos x (-2+\log x) +2x\csc x \log x )}{x^2}$ & $\frac{\cot^2 x \log^2 x }{x}$ \\ \hline
  \multicolumn{2}{c||}{} & $\frac{(\sec^2x)(9x+\sin(2x))(\tan^2(x))}{3\sqrt[3]{x}}$ & $\sqrt[3]{x}^2\tan^3 x$ \\ \hline 
  \multicolumn{2}{c||}{} & $\frac{n^2(\sec{x})(3+\cos (2x)-\sin(2x))\tan x )}{2e^x}$ & $\frac{n^2(\sin x)(\tan x)}{e^x}$ \\ \hline
  \end{tabular}
 \end{center}
\end{table}
}
\clearpage
{\renewcommand{\arraystretch}{2.5}
\begin{table}[b]
 \caption{Output results for each model for functions commonly incorrectly answered by learning model with polish.}
 \centering
 {\footnotesize
  \begin{tabular}{ll} \hline 
   Integrand & $\frac{-3}{2x^{\frac{5}{2}}}$ \\ \hline
   Primitive function & $x^{\frac{-3}{2}}$ \\ \hline
   Correct Answer (string) & power x divide minus 3 2 \\ \hline
   LSTM string polish output & divide 1 power x divide 3 2 \\ \hline
   Transformer string polish output & power x divide 3 2 \\ \hline
   Correct Answer (subtree) &  power x divide x EOS EOS divide -3 2 -3 EOS EOS 2 EOS EOS \\ \hline
   LSTM subtree polish output & divide 1 power 1 EOS EOS power x divide x EOS EOS divide 3 2 3 EOS EOS 2 EOS EOS \\ \hline
   Transformer subtree polish output & power 1 divide x EOS EOS divide 3 2 3 EOS EOS 2 EOS EOS \\ \hline \hline 

   Integrand & $-\frac{(2+\cos(2x))\csc^4 x}{n}$ \\ \hline
   Primitive function & $\frac{\cot x \csc^2x}{n}$ \\ \hline
   Correct Answer (string) & divide times cot x power csc x 2 n \\ \hline
   LSTM string polish output & divide times cot x csc x n \\ \hline
   Transformer string polish output & divide times cot x csc x n \\ \hline
   Correct Answer (subtree) & divide times n times cot power cot x EOS x EOS EOS power csc 2 csc x EOS \\
                            & x EOS EOS 2 EOS EOS n EOS EOS \\ \hline
   LSTM subtree polish output & divide times n times power power power cos 2 cos x EOS x EOS EOS 2 EOS EOS \\
                              & power csc 2 csc x EOS x EOS EOS 2 EOS EOS n EOS EOS \\ \hline
   Transformer subtree polish output & divide times n times power power power cot 2 cot x EOS x EOS EOS 2 EOS EOS \\ 
                                     & power csc 2 csc x EOS x EOS EOS 2 EOS EOS n EOS EOS \\ \hline
  \end{tabular}
 }
\end{table}
}
\clearpage
{\renewcommand{\arraystretch}{2.5}
{\footnotesize
\begin{longtable}{ll}
 \caption{Output results for each model for functions commonly answered incorrectly by learning model with Integrand Reverse polish Primitive Polish (IRPP).}
 \\
 \hline 
   Integrand & $\frac{\sec x(3x\sec x-2\sin x )\tan^2x}{x^3}$ \\ \hline
   Primitive function & $\frac{\tan^3 x}{x^2}$ \\ \hline
   Correct Answer (string) &  divide power tan x 3 power x 2 \\ \hline
   LSTM string IRPP output & divide times sec x power tan x 2 power x 2 \\ \hline
   Transformer string IRPP output & divide times sin x tan x power x 2 \\ \hline
   Correct Answer (subtree) & divide power power power tan 3 tan x EOS x EOS EOS 3 EOS EOS power x 2 \\
                            & x EOS EOS 2 EOS EOS \\ \hline
   LSTM subtree IRPP output & divide times power times sec power sec x EOS x EOS EOS power tan 2 tan x EOS \\
                            & x EOS EOS 2 EOS EOS power x 2 x EOS EOS 2 EOS EOS \\ \hline
   Transformer subtree IRPP output & divide power power power tan 2 tan x eos x eos eos 2 eos eos power x 3 \\  
                            & x eos eos 3 eos eos \\ \hline 
 \pagebreak    
 \hline 
   Integrand & $n\csc^2 x(-3\csc^2 x + \sec^2 x )$ \\ \hline
   Primitive function & $n\csc^3 x \sec x $ \\ \hline
   Correct Answer (string) & times times n power csc x 3 sec x \\ \hline
   LSTM string IRPP output & times times n power csc x 2 sec x \\ \hline
   Transformer string IRPP output & times times n power csc x 2 sec x \\ \hline
   Correct Answer (subtree) & times times sec times n power n EOS EOS power csc 3 csc x EOS x EOS EOS \\
                            & 3 EOS EOS sec x EOS x EOS EOS\\ \hline
   LSTM subtree IRPP output & times times power times n cot n EOS EOS cot x EOS x EOS EOS power csc 3 \\
                            & csc x EOS x EOS EOS 3 EOS EOS \\ \hline
   Transformer subtree IRPP output & times times sec times n power n EOS EOS power csc 2 csc x EOS x EOS EOS \\ 
                            & 2 EOS EOS sec x EOS x EOS EOS \\ \hline\hline
 
   Integrand & $\frac{1}{3\sqrt[3]{x}^2}$ \\ \hline
   Primitive function & $\sqrt[3]{x}$ \\ \hline
   Correct Answer (string) & root 3 x \\ \hline
   LSTM string IRPP output & power root 3 x 2 \\ \hline
   Transformer string IRPP output & power root 3 x minus 2 \\ \hline
   Correct Answer (subtree) & root 3 x 3 EOS EOS x EOS EOS \\ \hline
   LSTM subtree IRPP output & power root root root 3 x 3 EOS EOS x EOS EOS x EOS EOS \\ \hline
   Transformer subtree IRPP output & power root root root 3 x 3 EOS EOS x EOS EOS -1 EOS EOS \\ \hline    
   \pagebreak 
   \hline 
   Integrand & $\frac{n^2\sec x (3x+x\cos (2x)-\sin(2x))\tan x}{2x^2}$ \\ \hline
   Primitive function & $\frac{n^2\sin x \tan x}{x}$ \\ \hline
   Correct Answer (string) & divide times times power n 2 sin x tan x x \\ \hline
   LSTM string IRPP output & divide times times power n 2 sec x tan x x \\ \hline
   Transformer string IRPP output & divide times times power n 2 sec x tan x x \\ \hline
   Correct Answer (subtree) & divide times x times times tan times power sin power n 2 n EOS EOS 2 EOS EOS \\
                            & sin x EOS x EOS EOS tan x EOS x EOS EOS x EOS EOS \\ \hline
   LSTM subtree IRPP output & divide times x times times power times power sin power n 2 n EOS EOS 2 EOS EOS \\
                            & sin x EOS x EOS EOS power tan 2 tan x EOS x EOS EOS 2 EOS EOS x EOS EOS \\ \hline
   Transformer subtree IRPP output& divide times x times times tan times power sec power n 2 n EOS EOS 2 EOS EOS \\
                                  & sec x EOS x EOS EOS tan x EOS x EOS EOS x EOS EOS \\ \hline \hline
   Integrand & $-\frac {\cot x \csc x \log x (\cos x (-2+\log x) +2x\csc x \log x )}{x^2}$ \\ \hline
   Primitive function & $\frac{\cot^2 x \log^2 x }{x}$ \\ \hline
   Correct Answer (string) & divide times power cot x 2 power ln x 2 x \\ \hline
   LSTM string IRPP output & divide times times cos x power cot x 2 ln x x \\ \hline
   Transformer string IRPP output & divide times power cot x 2 csc x x \\ \hline
   Correct Answer (subtree) & divide times x times power power power cot 2 cot x EOS x EOS EOS 2 EOS EOS \\ 
                            & power ln 2 ln x EOS x EOS EOS 2 EOS EOS x EOS EOS \\ \hline
   LSTM subtree IRPP output & divide times x times cos power cos x EOS x EOS EOS power cot 2 cot x EOS \\
                            & x EOS EOS 2 EOS EOS x EOS EOS \\ \hline
   Transformer subtree IRPP output & divide times x times cos power cos x EOS x EOS EOS power cot 2 cot x EOS \\
                                   & x EOS EOS 2 EOS EOS x EOS EOS \\ \hline\hline 
   Integrand & $\frac{(\sec^2x)(9x+\sin(2x))(\tan^2(x))}{3\sqrt[3]{x}}$ \\ \hline 
   Primitive function & $\sqrt[3]{x}^2\tan^3 x$ \\ \hline
   Correct Answer (string) & times power root 3 x 2 power tan x 3 \\ \hline
   LSTM string IRPP output & times power root 3 x 2 power tan x 2 \\ \hline
   Transformer string IRPP output & times power sec x 3 power root 3 x 2 \\ \hline
   Correct Answer (subtree) & times power power power root 2 root 3 x 3 EOS EOS x EOS EOS 2 EOS EOS \\
                            & power tan 3 tan x EOS x EOS EOS 3 EOS EOS \\ \hline
   LSTM subtree IRPP output & times power power power tan 3 tan x EOS x EOS EOS 3 EOS EOS power root 2 \\ 
                            & root 3 x 3 EOS EOS x EOS EOS 2 EOS EOS \\ \hline
   Transformer subtree IRPP output & times power power power sec 3 sec x EOS x EOS EOS 3 EOS EOS power root 2 \\
                                   & root 3 x 3 EOS EOS x EOS EOS 2 EOS EOS \\ \hline
   \pagebreak 
   \hline 

   Integrand & $\frac{n^2(\sec{x})(3+\cos (2x)-\sin(2x))\tan x}{2e^x}$ \\ \hline
   Primitive function & $\frac{n^2(\sin x)(\tan x)}{e^x}$ \\ \hline
   Correct Answer (string) & divide times times power n 2 sin x tan x power e x \\ \hline
   LSTM string IRPP output & divide times times power n 2 sin x power tan x 2 power e x \\ \hline
   Transformer string IRPP output & divide times times n power sin x 2 tan x power e x \\ \hline
   Correct Answer (subtree) & divide times power times times tan times power sin power n 2 n EOS EOS 2 EOS EOS \\
                            & sin x EOS x EOS EOS tan x EOS x EOS EOS power e x e EOS EOS x EOS EOS \\ \hline
   LSTM subtree IRPP output & divide times power times times power times power sin power n 2 n EOS EOS 2 EOS EOS \\  
                            & sin x EOS x EOS EOS power tan 2 tan x EOS x EOS EOS 2 EOS EOS \\
                            & power e x e EOS EOS x EOS EOS \\ \hline
   Transformer subtree IRPP output & divide times power times times power times power sin power n 2 n EOS EOS 2 EOS EOS \\
                                   & sin x EOS x EOS EOS power tan 2 tan x EOS x EOS EOS 2 EOS EOS \\
                                   & power e x e EOS EOS x EOS EOS \\ \hline
\end{longtable}
}
}
\clearpage

\begin{table}[b]
 \caption{Tokens for operators, variables, and constants for string and subtree datasets.}
 \begin{center}
  \begin{tabular}{ccccccc} \hline 
   string token & subtree token & string token & \multicolumn{4}{c}{subtree token} \\  
   (Operator) & (Operator) & (Variable\&Constants) & \multicolumn{4}{c}{(Variable\&Constants)} \\ \hline 
   plus & plus & EOS & EOS & 10 &-3&-18\\ \hline 
   minus & minus & x & x & 11&-4&-19 \\ \hline 
   times & times & e & e & 12&-5&-21 \\ \hline 
   divide & divide & n & n & 13&-6&-24 \\ \hline 
   power & power & 0 & 0 & 14&-7& \\ \hline 
   root & root & 1 & 1 & 15&-8& \\ \hline 
   sqrt & sqrt & 2 & 2 & 17&-9& \\ \hline 
   sin & sin & 3 & 3 & 18&-10& \\ \hline 
   cos & cos & 4 & 4 & 19&-11& \\ \hline 
   tan & tan & 5 & 5 & 21&-12& \\ \hline 
   sec & sec & 6 & 6 & 23&-13& \\ \hline 
   csc & csc & 7 & 7 & 24&-14& \\ \hline 
   cot & cot & 8 & 8 & -1&-15& \\ \hline 
   ln & ln & 9 & 9 & -2&-17& \\ \hline 
  \end{tabular}
 \end{center}
\end{table} 

\clearpage
 \begin{table}[b]
 \caption{Hyperparameters of Long Short-Term Memory (LSTM) determined using the Tree-structured Parzen Estimator (TPE), one kind of Bayesian optimization.}
 \label{LSTM_hyperparameter}
  \centering
 \scalebox{0.8}{
 \begin{tabular}{ccccc}\hline 
  &LSTM string polish& LSTM subtree polish& LSTM string IRPP& LSTM subtree IRPP\\ \hline 
  Number of LSTM layer&3 &4 &5 &3 \\ \hline 
  Dimensionality&929 &384 &813 &1022 \\ \hline
  Dropout& 0.1396& 0.17721& 0.0404& 0.1974\\ \hline
  Batch size&128 &128 &128 &128 \\ \hline
  Gradient Clipping& 4.1231& 7.6506& 8.209& 4.532\\ \hline
  Optimizer& Adam($\alpha =0.0018$)& Adam($\alpha = 0.0005$)& Adam($\alpha = 0.00055$)& Adam($\alpha = 9.768\times 10^{-5}$) \\ \hline
  Weight Decay& $8.5900 \times 10^{-7}$ & $2.9503 \times 10^{-7}$ &0.000817 & $5.639\times 10^{-10}$\\ \hline
 \end{tabular}
 }
\end{table}
\clearpage
\bibliography{sci}
\bibliographystyle{naturemag}